\documentclass{article}

\usepackage{microtype}
\usepackage{graphicx}
\usepackage{subfigure}
\usepackage{booktabs} 

\usepackage{hyperref}

\usepackage{graphicx}
\usepackage{xcolor} 
\usepackage{bm}
\usepackage{multirow}
\usepackage{caption}
\usepackage{natbib}
\usepackage{wrapfig,lipsum,booktabs}

\usepackage{hyperref}
\usepackage{url}

\usepackage{graphicx}
\usepackage{xcolor} 
\usepackage{bm}
\usepackage{multirow}
\usepackage{caption}
\usepackage{natbib}
\usepackage{amsfonts}
\usepackage{amsmath}

\usepackage{algorithm}
\usepackage{algpseudocode}

\usepackage{sidecap, caption}
\usepackage{hyperref}

\usepackage[accepted]{icml2023}

\usepackage{amsmath,amsfonts,bm}

\def\eqref#1{equation~\ref{#1}}

\def\1{\bm{1}}

\DeclareMathAlphabet{\mathsfit}{\encodingdefault}{\sfdefault}{m}{sl}
\SetMathAlphabet{\mathsfit}{bold}{\encodingdefault}{\sfdefault}{bx}{n}

\newcommand{\E}{\mathbb{E}}

\usepackage{amsmath}
\usepackage{amssymb}
\usepackage{mathtools}
\usepackage{amsthm}

\usepackage[capitalize,noabbrev]{cleveref}

\theoremstyle{plain}

\theoremstyle{definition}

\theoremstyle{remark}

\usepackage[textsize=tiny]{todonotes}

\icmltitlerunning{GFlowOut: Dropout with Generative Flow Networks}

\begin{document}

\twocolumn[
\icmltitle{GFlowOut: Dropout with Generative Flow Networks}




\begin{icmlauthorlist}
\icmlauthor{Dianbo Liu}{mila,broad}
\icmlauthor{Moksh Jain}{mila,udem}
\icmlauthor{Bonaventure F. P. Dossou}{mila,mcgill,lelapa}
\icmlauthor{Qianli Shen}{nus}
\icmlauthor{Salem Lahlou}{mila,udem}
\icmlauthor{Anirudh Goyal}{deepmind}
\icmlauthor{Nikolay Malkin}{mila,udem}
\icmlauthor{Chris C. Emezue}{mila,tum}
\icmlauthor{Dinghuai Zhang}{mila,udem}

\icmlauthor{Nadhir Hassen}{mila,udem}
\icmlauthor{Xu Ji}{mila,udem}
\icmlauthor{Kenji Kawaguchi}{nus}
\icmlauthor{Yoshua Bengio}{mila,udem,cifar}

\end{icmlauthorlist}

\icmlaffiliation{mila}{Mila Quebec AI Institute}
\icmlaffiliation{broad}{Broad Institute of MIT and Harvard}
\icmlaffiliation{udem}{University of Montreal}
\icmlaffiliation{mcgill}{McGill University}
\icmlaffiliation{lelapa}{Lelapa AI}
\icmlaffiliation{nus}{National University of Singapore}
\icmlaffiliation{deepmind}{Google DeepMind}
\icmlaffiliation{tum}{Technical University of Munich}
\icmlaffiliation{cifar}{CIFAR AI Chair}

\icmlcorrespondingauthor{Dianbo Liu}{dianbo.liu@mila.quebec}
 \icmlcorrespondingauthor{Moksh Jain}{moksh.jain@mila.quebec}

\icmlkeywords{Machine Learning, ICML}

\vskip 0.3in
]



\printAffiliationsAndNotice{} 

\begin{abstract}
Bayesian inference offers principled tools to tackle many critical problems with modern neural networks such as poor calibration and generalization, and data inefficiency. However, scaling Bayesian inference to large architectures is challenging and requires restrictive approximations. Monte Carlo Dropout has been widely used as a relatively cheap way to approximate inference and estimate uncertainty with deep neural networks. Traditionally, the dropout mask is sampled independently from a fixed distribution. Recent research shows that the dropout mask can be seen as a latent variable, which can be inferred with variational inference. These methods face two important challenges: (a) the posterior distribution over masks can be highly multi-modal which can be difficult to approximate with standard variational inference and (b) it is not trivial to fully utilize sample-dependent information and correlation among dropout masks to improve posterior estimation. In this work, we propose GFlowOut to address these issues. GFlowOut leverages the recently proposed probabilistic framework of Generative Flow Networks (GFlowNets) to learn the posterior distribution over dropout masks. We empirically demonstrate that GFlowOut results in predictive distributions that generalize better to out-of-distribution data and provide uncertainty estimates which lead to better performance in downstream tasks.
\end{abstract}

\section{Introduction}
A key shortcoming of modern deep neural networks is that they are often overconfident about their predictions, especially when there is a distributional shift between train and test dataset~\cite{daxberger2021bayesian,nguyen2015deep,guo2017calibration}. In risk-sensitive scenarios such as clinical practice and drug discovery, where mistakes can be extremely costly, it is important that models provide predictions with reliable uncertainty estimates~\cite{bhatt2021uncertainty}. Bayesian inference offers principled tools to model the parameters of neural networks as random variables, placing a prior on them and inferring their posterior given some observed data~\cite{mackay1992practical,neal2012bayesian}. The posterior captures the uncertainty in the predictions of the model and also serves as an effective regularization strategy resulting in improved generalization~\cite{wilson2020bayesian,lotfi2022bayesian}. In practice, exact Bayesian inference is often intractable and existing Bayesian deep learning methods rely on assumptions that result in posteriors that are less expressive and can provide poorly calibrated uncertainty estimates~\cite{ovadia2019can,fort2019deep,foong2020expressiveness,daxberger2021bayesian}. In addition, even with several approximations, Bayesian deep learning methods are often significantly more computationally expensive and slower to train compared to non-Bayesian methods~\cite{kuleshov2018accurate,boluki2020learnable}.

\begin{SCfigure*}
  \vspace{-2mm}
  \centering
  \includegraphics[width=0.90\linewidth]{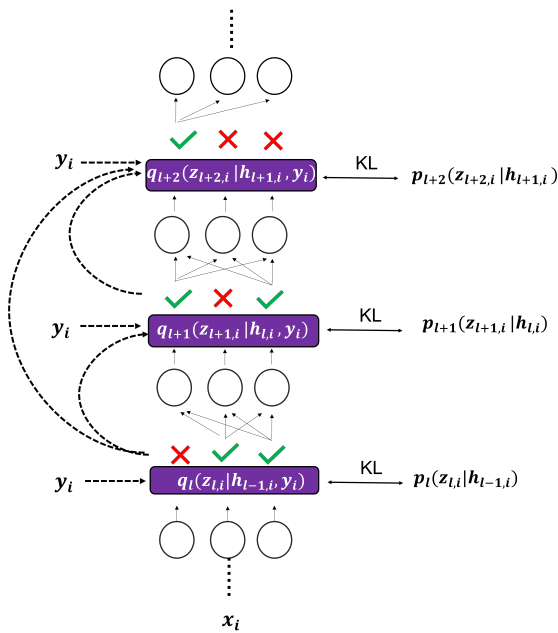}
  \caption{In this work, we propose a Generative Flow Network (GFlowNet) based binary dropout mask generator which we refer to as \textbf{GFlowOut}. Purple squares are GFlowNet-based dropout mask generators parameterized as multi-layer perceptrons. $z_{i,l}$ refers to dropout masks for data point indexed by $i$ at layer $l$ of the model. $h_{i,l}$ refers to activations of the model at layer $l$ given input $x_i$.  $q(\cdot)$ are  auxiliary  variational functions used and adapted only during model training, in which the posterior distribution over dropout masks is conditioned implicitly on input covariates ($x_i$) and directly on the label ($y_i$) of the data point to make the estimation easier. $p(\cdot)$ are   mask generation functions used at test time, which are only conditioned on $x_i$ and trained by minimizing the Kullback–Leibler(KL) divergence with $q(\cdot)$. In addition, both  $q(\cdot)$ and $p(\cdot)$ conditions explicitly on dropout masks of all previous layers.
}
\label{fig:GFlowOut}
\vspace{-2mm}
\end{SCfigure*}
\label{method}

\citet*{gal2016dropout} show that deep neural networks with dropout perform approximate Bayesian inference and approximate the posterior of a deep Gaussian process~\cite{damianou2013deep}. One can obtain samples from this predictive distribution by taking multiple forward passes through the neural network with independently sampled dropout masks. Due to its simplicity and minimal computational overhead, dropout has since been used as a method to estimate uncertainty and improve robustness in neural networks. Different variants of dropout have been proposed and can be interpreted as different variational approximations to model the posterior over the neural network parameters~\cite{ba2013adaptive,kingma2015variational,gal2017concrete,ghiasi2018dropblock,fan2021contextual,pham2021autodropout}. 

There are a few major challenges in approximating the Bayesian posterior over model parameters using dropout: (1) the multimodal nature of the posterior distribution makes it difficult to approximate with standard variational inference~\cite{gal2016dropout,le2021mc}, which assumes factorized priors; (2) dropout masks are discrete objects making gradient-based optimization difficult~\cite{boluki2020learnable}; (3) variational inference methods can suffer from high gradient variance resulting in optimization instability~\cite{kingma2015variational}; (4) modeling dependence between dropout masks from different layers is non-trivial. 

The recently proposed Generative Flow Networks (GFlowNets)~\cite{bengio2021flow,bengio2021gflownet} frame the problem of generating discrete objects as a control problem based on the sequential construction of discrete components. GFlowNets learn probabilistic policies that sample objects proportional to a reward function (or exp(-energy)). They have demonstrated better generalization to multimodal distributions~\cite{nica2022evaluating} and have lower gradient variance compared with policy gradient-based variational methods \cite{malkin2022gflownets}, making it an interesting choice for posterior inference for dropout.

\textbf{Contributions.} In this work, to address the limitations of standard variational inference, we develop a GFlowNet-based binary dropout mask generator which we refer to as \textit{GFlowOut}, to estimate the posterior distribution of binary dropout masks. GFlowOut generates dropout masks for a layer, conditioned on masks generated for the previous layer, therefore accounting for inter-layer dropout dependence. Furthermore, the GFlowOut estimator can be conditioned on the data point: GFlowOut improves posterior estimation here by utilizing both input covariates and labels in the training set of supervised learning tasks via an auxiliary variational function. To investigate the quality of the posterior distribution learned by GFlowOut, we design empirical experiments, including evaluating robustness to distribution shift during inference, detecting out-of-distribution examples with uncertainty estimates, and transfer learning, using both benchmark datasets and a real-world clinical dataset.

\section{Related work}

\subsection{Dropout as a Bayesian approximation}
Deep learning tools have shown tremendous power in different applications. However, traditional deep learning tools lack mechanisms to capture the uncertainty, which is of crucial importance in many fields.
Uncertainty quantification (UQ) is studied extensively as a fundamental problem of deep learning and a large number of Bayesian deep learning tools have emerged in recent years.
For example, \citet{gal2016dropout} showed that casting dropout in deep learning model training is an approximation of Bayesian inference in deep Gaussian processes and allows uncertainty estimation without extra computational cost. 
\citet{kingma2015variational} proposed variational dropout, where a dropout posterior over parameters is learned by treating dropout regularization as approximate inference in deep models.
\citet{gal2017concrete} developed a continuous relaxation of discrete dropout masks to improve uncertainty estimation, especially in reinforcement learning settings. \citet{lee2020meta} introduced ``meta-dropout'', which involves an additional global term shared across all data points during inference to improve generalization. \citet{xie2019soft} replaced the hard dropout mask following a Bernoulli distribution with the soft mask following a beta distribution and conducted the optimization using a stochastic gradient variational Bayesian algorithm to control the dropout rate.  \citet{boluki2020learnable} combined a model-agnostic dropout scheme with variational auto-encoders (VAEs), resulting in semi-implicit VAE models. 
Instead of using mean-field family for variational inference, \citet{nguyen2021structured} utilized a structured representation of multiplicative Gaussian noise for better posterior estimation. 
More recently, \citet{fan2021contextual} developed ``contextual dropout'', which optimizes variational objectives in a sample-dependent manner and, to the best of our knowledge, is the closest approach to GFlowOut in the literature. 
GFlowOut differs from contextual dropout in several aspects. 
First, both methods take trainable priors into account, but GFlowNet also takes into account priors that depend on the input covariate of each data point.
Second, the variational posterior of contextual dropout only depends on the input covariate ($x$), while in GFlowOut, the variational posterior is also conditioned on the label $y$, which provides more information for training. 
Third, within each neural network layer, the mask of contextual dropout is conditioned on previous masks implicitly, while the mask of GFlowOut is conditioned on previous masks explicitly by directly feeding previous masks as inputs into the generator, which improves the training process.
Finally, instead of a REINFORCE-based gradient estimator used for contextual dropout training, GFlowOut employs powerful GFlowNets for the variational posterior.

\vspace{-1mm}
\subsection{Generative flow networks}
\vspace{-1mm}
Generative flow networks (GFlowNets)~\cite{bengio2021flow,bengio2021gflownet} are a family of probabilistic models that amortizes sampling discrete compositional objects proportionally to a given unnormalized density function. GFlowNets learn a stochastic policy to construct objects through a sequence of actions akin to deep reinforcement learning~\cite{sutton2018reinforcement}. GFlowNets are trained so as to make the likelihood of reaching a terminating state proportional to the reward. Recent works have shown close connections of GFlowNets to other generative models~\cite{zhang2022unifying} and to hierarchical variational inference~\cite{malkin2022gflownets}. GFlowNets achieved great empirical success in learning energy-based models~\cite{zhang2022generative}, small-molecule generation~\cite{bengio2021flow,nica2022evaluating,malkin2022trajectory,madan2022learning,Pan2022GenerativeAF}, biological sequence generation~\cite{malkin2022trajectory,jain2022biological,madan2022learning}, and structure learning~\cite{deleu2022bayesian}. 
Several training objectives have been proposed for GFlowNets, including Flow Matching (FM)~\cite{bengio2021flow}, Detailed Balance (DB)~\cite{bengio2021gflownet}, Trajectory Balance (TB)~\cite{malkin2022trajectory}, and the more recent Sub-Trajectory Balance (SubTB)~\cite{madan2022learning}. In this work, we use the Trajectory Balance (TB) objective.


\section{Method}

In this section, we define the problem setting and mathematical notations used in this study, as well as describe the proposed method, GFlowOut, for dropout mask generation in detail.

\subsection{Background and notation}
\textbf{Dropout.} In a vanilla feed-forward neural network (MLP) with $L$ layers, each layer of the model has weight matrix $w_l$ and bias vector $b_l$. It takes as input activations $h_{l-1}$ from previous layer with layer index $l-1$, and computes as output $h_{l} = \sigma(w_l h_{l-1} + b_l)$ where $\sigma$ is a non-linear activation function. Dropout consists of dropping out units from the output of a layer. Formally this can be described as applying a sampled binary mask $z_l\sim p(z_l)$ on the output of the layer $h_{l} = z_l \circ \sigma(w_l h_{l-1} + b_l)$, at each layer in the model. In regular random dropout, $z_l$ is a collection of i.i.d.\ ${\rm Bernoulli}(r)$ variables, where $r$ is a fixed parameter for all the layers. Recently, several approaches have been proposed to learn $p(z_l)$ along with the model parameters. In these approaches, $z$ is viewed either as latent variables or part of the model parameters. We consider two variants for our proposed method: GFlowOut where the dropout masks $z$ are viewed as \textit{sample dependent} latent variables, and  ID-GFlowOut, which generates masks in a \textit{sample independent} manner where $z$ is viewed as a part of the model parameters shared across all samples. Next, we briefly introduce GFlowNets and describe how they model the dropout masks $z$ given the data.


\textbf{GFlowNets.} Let $G=(\mathcal{S}, \mathcal{A})$ be a directed acyclic graph (DAG) where the vertices $s\in\mathcal{S}$ are \emph{states}, including a special initial state $s_0$ with no incoming edges, and directed edges $(s\rightarrow s')\in \mathcal{A}$ are \emph{actions}. $\mathcal{X}\subseteq \mathcal{S}$ denotes the terminal states, with no outgoing edges. A complete trajectory $\tau = (s_0\rightarrow \dots s_{i-1}\rightarrow s_i \dots \rightarrow z) \in \mathcal{T}$ in $G$ is a sequence of states starting at $s_0$ and terminating at $z\in \mathcal{X}$ where each $(s_{i-1}\rightarrow s_i)\in \mathcal{A}$. The forward policy $P_F(-|s)$ is a collection of distributions over the children of each non-terminal node $s\in\mathcal{S}$ and defines a distribution over complete trajectories, $P_F(\tau) = \prod_{(s_{i-1}\rightarrow s_i) \in\tau}P_F(s_i|s_{i-1})$. We can sample terminal states $z\in\mathcal{X}$ by sampling trajectories following $P_F$. Let $\pi(x)$ be the marginal likelihood of sampling terminal state $x$, $\pi(z) = \sum_{\tau=(s_0\rightarrow\dots \rightarrow z)\in \mathcal{T}}P_F(\tau)$. Given a non-negative reward function $R: \mathcal{X} \to \mathbb{R}^+$, the learning problem tackled in GFlowNets is to estimate $P_F$ such that $\pi(z)\propto R(z),\enspace \forall z\in \mathcal{X}$. We refer the reader to~\citet{bengio2021gflownet,malkin2022trajectory} for a more thorough introduction to GFlowNets.

We adopt the Trajectory Balance (TB)~\cite{malkin2022trajectory} parameterization, which includes $P_F(-|-;\phi), P_B(-|-; \phi)$, and $Z_\gamma$, where $\phi$ and $\gamma$ are the learnable parameters. The backward policy $P_B$ is a distribution over parents of every noninitial state, and $Z$ is an estimate of the partition function. 

Within the context of generating dropout masks, a complete dropout mask $x\in \mathcal{X}$ is a binary vector of dimension $M$, where $M$ is the number of units in the neural network, i.e. $\mathcal{X}$ is equal to $\{0, 1\}^M$. A partially constructed mask $s\in\mathcal{S}$ is a binary vector of dimension $m < M$ representing the mask for a set of initial layers in the model, and an action consists of appending the mask for the subsequent layer to this vector. That is, each action in the sequence samples the mask for an entire layer (in parallel), conditioned on the masks for the previous layers.

In the next section, we formally describe how GFlowNets can be used for generating dropout masks, as well as practical implementation details.


\begin{algorithm}[]
\caption{GFlowOut}\label{alg:GFlowOut}
\begin{algorithmic}
\State The whole system has the following 3 components:
\begin{itemize}
    \item \emph{Backbone Model} (eg, classifier) neural network $p(y_i|x_i,z_i;\theta)$.This algorithm section is written assuming $p(y_i|x_i,z_i;\theta)$ is an MLP with $L$ hidden layers, but it can be easily extended to other architectures
    
    \item \emph{GFlowNet} $q(z_i|x_i,y_i;\phi)$ which approximates the posterior distribution over dropout masks $z_i$ conditioned on both $x_i$ and $y_i$ from the data point. Its tempered version $q^{\sim}(z_i|x_i,y_i;\phi)$ is used for dropout mask sampling during training.
    
    \item $p(z_i|x_i;\xi)$, which generates dropout mask distribution only conditioned on $x_i$, is optimized by minimizing KL divergence with $q(z_i|x_i,y_i;\phi)$  and is used for dropout mask sampling during test time.
  
\end{itemize}

\vspace{0.5cm}
\State $q(z_i|x_i,y_i;\phi)$ and $p(z_i|x_i;\xi)$  are all implemented as groups of MLPs, one MLP for each layer $l$ and they do not share parameters with each other nor between different layers. 

\State Next, we explain how dropout masks are generated and how $p(y_i|x_i,z_i;\theta)$, $q(z_i|x_i,y_i;\phi)$ and $p(z_i|x_i;\xi)$  are computed.

\vspace{0.5cm}

\For{epoch}
\For{Iterate data point $x_i,y_i$} \Comment{(batches used in actual training)}

    \State $h_0=x_i$
    \vspace{0.2cm}
    \State$var1=0$\Comment{Variables to store probabilities from each layer}
    \State $var2=0$

    \For{ layer $l$ in $1:L-1$}
        \State \textbf{Use current layer's activation and dropout masks of all previous layers for mask generation} 

        \State $h'_{l,i} = {\rm ReLU}(b_l+w_l h_{l-1,i})$
         
        \State $z_{i,l} \sim q^{\sim}_l(z_{i,l}|h'_{l,i},y_i,(z_{i,j})^{l-1}_{j=1};\phi)$  \Comment{dropout masks generated }
        \State $h_{l,i} = z_{i,l}h'_{l,i}$ \Comment{apply dropout }
        
        \State$var1+=\log q_l(z_{i,l}|x_i,y_i,(z_{i,j})^{l-1}_{j=1};\phi)$
        \Comment{calculate log probabilities }
        \State $var2 += \log p_l(z_{i,l}|x_i(z_{i,j})^{l-1}_{j=1};\xi)$

    \EndFor
    
    \State$\log q(z_{i}|x_i,y_i;\phi)=var1$
    \State $\log p(z_{i}|x_i;\xi) =var2$
  
    \vspace{0.5cm}
    \State In the output layer, $\hat{y_i} = f_{\rm out}(b_L + w_L h_{L-1,i})$
    \State where $f_{out}$ is the output non-linearity of the output layer

    \State Update $\theta,\phi,\xi,\omega,\gamma$ using equations 4-12
    
\EndFor
\EndFor
\end{algorithmic}
\end{algorithm}


\subsection{GFlowOut}

We consider a generative model of the form $p(x, y, z) = p(x) p(z | x) p(y | x, z)$, where $x$ is the input data with corresponding label $y$ and $z$ is a local discrete latent variable representing the sample-dependent dropout mask, along with a dataset of observations $D=\{(x_i, y_i)\}_{i=1}^N$. GFlowOut learns to approximate the posterior $p(z | x, y)$ using the given dataset $D$. 
In a supervised learning task where the goal is to learn the predictive distribution $p(y|x)$, with the assumed generative model above, the following variational bound can be derived:
\begin{align}
    &\log \prod^N_{i=1}p(y_i|x_i)  \\
    &=\log \prod^N_{i=1} \sum_{z_i \in \mathcal{X}}p(z_i|x_i) p(y_i|x_i,z_i) \nonumber \\
    &= \log \prod^N_{i=1}
    \sum_{z_i \in \mathcal{X}}p(z_i|x_i)\frac{q(z_i|x_i,y_i)}{q(z_i|x_i,y_i)}p(y_i|x_i,z_i) \nonumber \\
   &= \sum_{i=1}^N \log  \mathop{\mathbb{E}}_{q(z_i|x_i,y_i)}\left[ \frac{p(z_i|x_i)}{q(z_i|x_i,y_i)}p(y_i|x_i,z_i)\right]  \\
   &\geq \sum_{i=1}^N \E_{q(z_i|x_i,y_i)}\left[ \log \frac{p(z_i|x_i)}{q(z_i|x_i,y_i)}p(y_i|x_i,z_i)\right] \nonumber\\
   &= \sum_{i=1}^N \bigg[ \E_{q(z_i|x_i,y_i)}[\log p(y_i|x_i,z_i)] \nonumber \\ & \quad\quad\quad - {\rm KL}(q(z_i|x_i,y_i)\|p(z_i|x_i))\bigg]
\end{align}
where $p(z_i|x_i)$ is part of the generative process and $q(z_i|x_i,y_i)$ is the variational distribution used to approximate the posterior of $z_i$. To improve the efficiency of training and fully utilize the information available in each data point, we design  $q(z_i|x_i,y_i)$ so that the distribution of $z_i$ is conditioned on both $x_i$ and $y_i$. As a consequence, $q(z_i|x_i,y_i)$ is not accessible during inference where $y_i$ is not available. Instead, $p(z_i|x_i)$, which is trained by minimizing KL divergence with $q(z_i|x_i,y_i)$, is used for inference.







Parametrizing each of the terms as $p(y_i|x_i,z_i;\theta)$, $q(z_i|x_i,y_i;\phi)$ and  $p(z_i|x_i;\xi)$, the goal is to maximize the lower bound derived above, which we denote as:
%
\begin{align}\mathcal{B}(\mathcal{D}; \theta, \phi, \xi)=\sum^N_{i=1}\bigg[\mathop{\mathbb{E}}_{q(z_i|x_i,y_i;\phi)}[ \log p(y_i|x_i,z_i;\theta)] \nonumber\\
\quad\quad-{\rm KL}(q(z_i|x_i,y_i;\phi)\|p(z_i|x_i;\xi))\bigg] 
\end{align}
%


The gradients of $\mathcal{B}$ with respect to its parameters are:
\begin{align*}
    \nabla_{\theta} \mathcal{B}(\mathcal{D}; \theta, \phi, \xi)&= \sum_{i=1}^N \nabla_\theta \E_{q(z_i|x_i,y_i)}[\log p(y_i|x_i,z_i; \theta)] \\
    \nabla_{\xi} \mathcal{B}(\mathcal{D}; \theta, \phi, \xi) &= \sum_{i=1}^N \nabla_\xi \E_{q(z_i|x_i,y_i)}[\log p(z_i \mid x_i; \xi)] 
\end{align*}






The gradient of the variational objective $\mathcal{B}$ with respect to $\phi$ requires a score function estimator, which is known to suffer from high gradient variance ~\cite{malkin2022gflownets}. Instead of directly optimizing $\mathcal{B}$ with respect to $\phi$, we first observe that $\mathcal{B}$ can be written as:
\begin{align*}
    \mathcal{B}(\mathcal{D}) = \sum_{i=1}^N \left(\log p (y_i | x_i) - {\rm KL}(q(z_i|x_i, y_i) \| p(z_i|x_i, y_i)) \right),
\end{align*}
making $p(z_i|x_i, y_i)$, the true posterior, a target for the variational distribution $q(z_i|x_i, y_i)$. We thus
propose to use a GFlowNet with the Trajectory Balance loss to train $q(z_i|x_i,y_i;\phi)$ to match its target, given by its unnormalized density $R=p(y_i | x_i, z_i) p (z_i | x_i)$. As binary dropout masks $z$ are high-dimensional discrete objects that can be constructed sequentially, we consider them as the terminating states of a GFlowNet, and instead of learning a distribution over these terminating states directly, we exploit the DAG structure to learn a forward policy $P_F(-|-;\phi)$, for which the terminating state distribution is $q(z_i|x_i,y_i;\phi)$. The Trajectory Balance loss  requires an additional parameter $Z_\gamma$, to train $q(z_i|x_i,y_i;\phi)$ (\cite{bengio2021gflownet,malkin2022trajectory}).

Corresponding trajectory balance loss for a trajectory $\tau=(s_0,...s_L)$ w.r.t.  $(Z_\gamma, P_F(-|-;\phi))$ will be
\begin{align}
\mathcal{L}_{TB}(\tau, \mathcal{D}; \phi,\gamma) = \left(\log \frac{Z_\gamma\prod_{t=1}^{L} P_F(s_t|s_{t-1};\phi)}{R}\right)^2
\end{align}
where a state $s_l$ in the GFlowNet graph refers to the set of dropout masks sampled by the GFlowNet from layer 1 to $l$ of the model ($(z_j)^l_{j=1}$). $L$ is the number of layers involving dropout. $s_L$ indicates the termination of the trajectory sampling process. $P_F(s_t|s_{t-1};\phi)$ refers to the forward policy in GFlowNets \footnote{As the graph $G$ used in this study has a tree structure, the backward policy $P_B(s_{t-1}|s_t)$ is a constant so we leave it out of the equations.}. $\log Z_\gamma=f(x_i,y_i;\gamma)$ is the partition function estimator with parameters $\gamma$ conditioned on both $x_i$ and $y_i$ from the data point indexed by $i$. Its parameter $\gamma$ is trained together with $\phi$. $R$ is the reward calculated from the likelihood of the data and the prior distribution of the states which are sets of  dropout masks (see equation 12).

The parameters $\phi$ and $\gamma$ are updated by taking gradient steps on $\mathcal{L}_{TB}(\tau, \mathcal{D}; \phi,\gamma)$ for $\tau$ sampled from some training policy. We choose to make the training policy a tempered version of $q(z_i|x_i,y_i;\phi)$, denoted  $q^{\sim}(z_i|x_i,y_i;\phi)$. The expected gradient update is thus equal to 
\begin{align}  
\sum_{i=1}^N \mathop{\mathbb{E}}_{z_i \sim q^{\sim}(\cdot|x_i,y_i;\phi)} \nabla_{\phi,\gamma} (&\log Z_\gamma +\log q(z_i|x_i,y_i;\phi)\nonumber \\ &-\log R)^2
\end{align}
\vspace{-2mm}
where
$$
\log R =\log p(y_i|x_i,z_i;\theta)+ \log p(z_i|x_i,\xi).
$$
During inference one estimates the posterior predictive as $p^{pred}(y_i|x_i)=\frac{1}{M}\sum^M_{j=1} p(y_i|x_i,z_j;\theta)$ where $M$ different $z_j$ are sampled from the $p(z_i|x_i;\xi)$ distribution.

\textbf{Implementation details.} 
Algorithm~\ref{alg:GFlowOut} 
 presents a high-level overview of the GFlowOut implementation. $q(z_i|x_i,y_i;\phi)$ is implemented as a set of multiple MLPs, one for each layer in the model that requires dropout mask generation (see Figure~\ref{fig:GFlowOut}). At layer $l$ of the model, the dropout probabilities of all units in layer $l$ are estimated in parallel conditioned on previous layer's activation $h_{l-1,i}$, the label $y_i$, and all dropout mask in layers before $l$ using $q_l(z_{i,l}|h_{l-1,i},y_i,(z_{i,j})^{l-1}_{j=1};\phi)$ which is parameterized as an MLP. The same process is repeated for each layer in the model. In this way, dropout mask probability at layer $l$ takes into consideration  input $x_i$ through $h_{l-1, i}$, label $y_i$ and joint probability with all dropout mask in previous layers, but are independent of masks in the same layer. $p(z_i|x_i;\xi)$ follows the same implementation except that it is not conditioned on $y_i$ and hence it can be used at test time for prediction. In convolutional neural networks, GFlowOut is implemented in a similar manner except that the units are dropped out channel-wise~\cite{yang2020rx,park2016analysis}. Early stopping based on performance on the validation set is used to prevent overfitting. Details of the computational efficiency of GFlowOut are discussed in the Appendix.

\subsection{ID-GFlowOut: GFlowOut without sample-dependent information}
\vspace{-2mm}
To understand if the sample-dependent information is needed, we introduce a variant of GFlowOut that only uses sample-independent information and keeps the rest of the algorithm as close to GFlowOut as possible for comparison. 

Consider a generative model of the form $p(x, y, z) = p(x) p(z) p(y | x, z)$. Given a supervised learning task $p(y|x)$, we generate a dropout mask $z$ that is \textit{not} conditioned on the data point. We use $q(z)$ to approximate the posterior of $z$ which can be seen as part of the model parameters shared by all data points. The following equations can be derived:
\begin{align}
\log \prod^N_{i=1}p(y_i|x_i)=\log \sum_{z}p(z)\prod^N_{i=1}  p(y_i|x_i,z) \nonumber \\
=\log \sum_{z}p(z)\frac{q(z)}{q(z)}\prod^N_{i=1}  p(y_i|x_i,z) \nonumber\\
=\log \mathop{\mathbb{E}}_{z \sim q(z)}[ \frac{p(z)}{q(z)} \prod^N_{i=1}p(y_i|x_i,z)] \nonumber \\
\geq  \mathop{\mathbb{E}}_{z \sim q(z)}\log \left[ \frac{p(z)}{q(z)} \prod^N_{i=1} p(y_i|x_i,z)\right] \nonumber \\
=\mathop{\mathbb{E}}_{z \sim q(z)}\left[\sum^N_{i=1} \log p(y_i|x_i,z)\right]-{\rm KL}(q(z)\|p(z))  
\end{align}
where the same distribution $p(z)$ is shared across the whole dataset and $q$ is conditional on the whole data set implicitly.
\begin{align}\mathcal{B}(\mathcal{D}; \theta', \phi' )=\mathop{\mathbb{E}}_{q(z;\phi')}\sum^N_{i=1} \log p(y_i|x_i,z;\theta') \nonumber\\
-{\rm KL}(q(z;\phi')||p(z) )
\end{align}
Where $\mathcal{B}$ is a lower bound. We parameterized each of the terms as $p(y|x,z;\theta')$ and $q(z;\phi')$. $p(z)$ is set as fixed prior to each unit of the dropout rate of 0.5. The gradients for stochastic optimization of $\theta$ can be obtained as
\begin{align}
\nabla_{\theta} \mathcal{B}(\mathcal{D}; \theta', \phi', \xi')= \sum_{i=1}^N \mathop{\mathbb{E}}_{z \sim q(z;\phi')} \nabla_{\theta'}  \log p(y_i|x_i,z;\theta')
\end{align}
The distribution $q(z;\phi')$ can be trained as a the policy of a GFlowNet, using a tempered version $q^{\sim }(z;\phi')$ to sample trajectories for training. The expected update direction for $\phi'$ and $\gamma'$ can be shown to equal
\begin{align} 
\sum_{i=1}^N \mathop{\mathbb{E}}_{z_i \sim q^{\sim}(z_i;\phi')} \nabla_{\phi,\gamma'} (\log Z_{\gamma'}+ \log q(z_i ;\phi')-\log R)^2,
\end{align}
where the log-reward for is
\begin{align}
\log R =N\log p(y_i|x_i,z_i;\theta')+ \log p(z_i).
\end{align}
In ID-GFlowOut, $\log Z_{\gamma'}=\gamma'$ and does not condition on any inputs.

During inference, the posterior predictive estimate is $p^{pred}(y_i|x_i)=\frac{1}{M}\sum^M_{j=1} p(y_i|x_i,z_j;\theta')$ where $M$ different $z_j$ are sampled from the $q(z;\phi')$ distribution.


\section{Experiments}

In this section, we empirically evaluate GFlowOut\footnote{Code is available at \url{https://github.com/kaiyuanmifen/GFNDropout}} on a variety of tasks to understand its ability to generalize across different distributions and estimate uncertainties in prediction. We first evaluate the generalization performance of the posterior predictive approximated by GFlowOut on an image classification task. We also evaluate the efficacy of GFlowOut in the context of transfer learning.  To understand the performance of GFlowOut when used in larger models and datasets, we conduct Visual Question Answering (VQA) experiments using Transformer architectures. Next, we evaluate the uncertainty captured by the posterior to detect out-of-distribution examples.  Finally, we study a potential application of GFlowOut in a real-world clinical use case for the cross-hospital prediction of mortality in intensive care units (ICUs). We supplement these results with further analysis and additional experimental details in the Appendix.

\begin{figure*}[]
    \vspace{-2mm}
    \centering
    \includegraphics[width=0.40\linewidth]{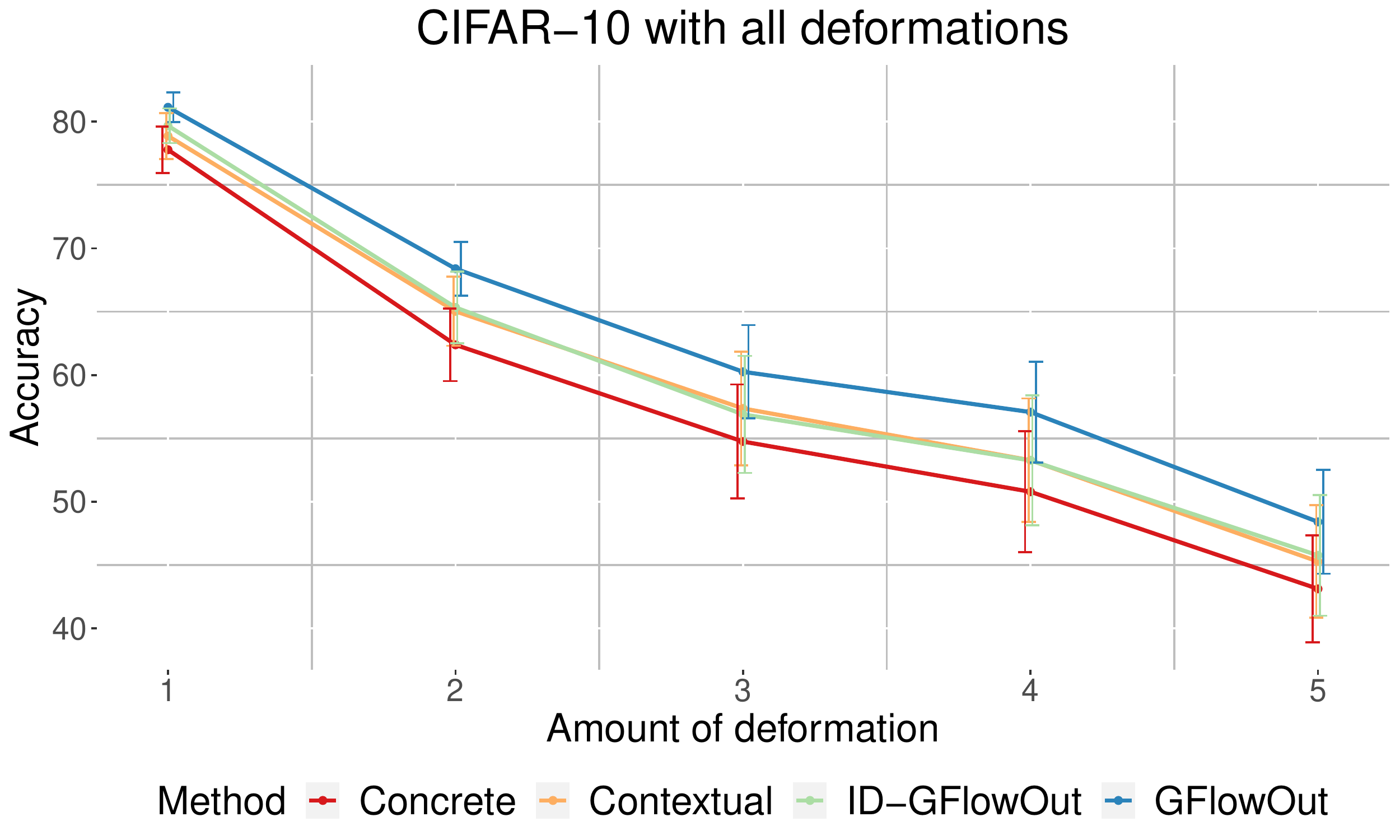}
    \includegraphics[width=0.40\linewidth]{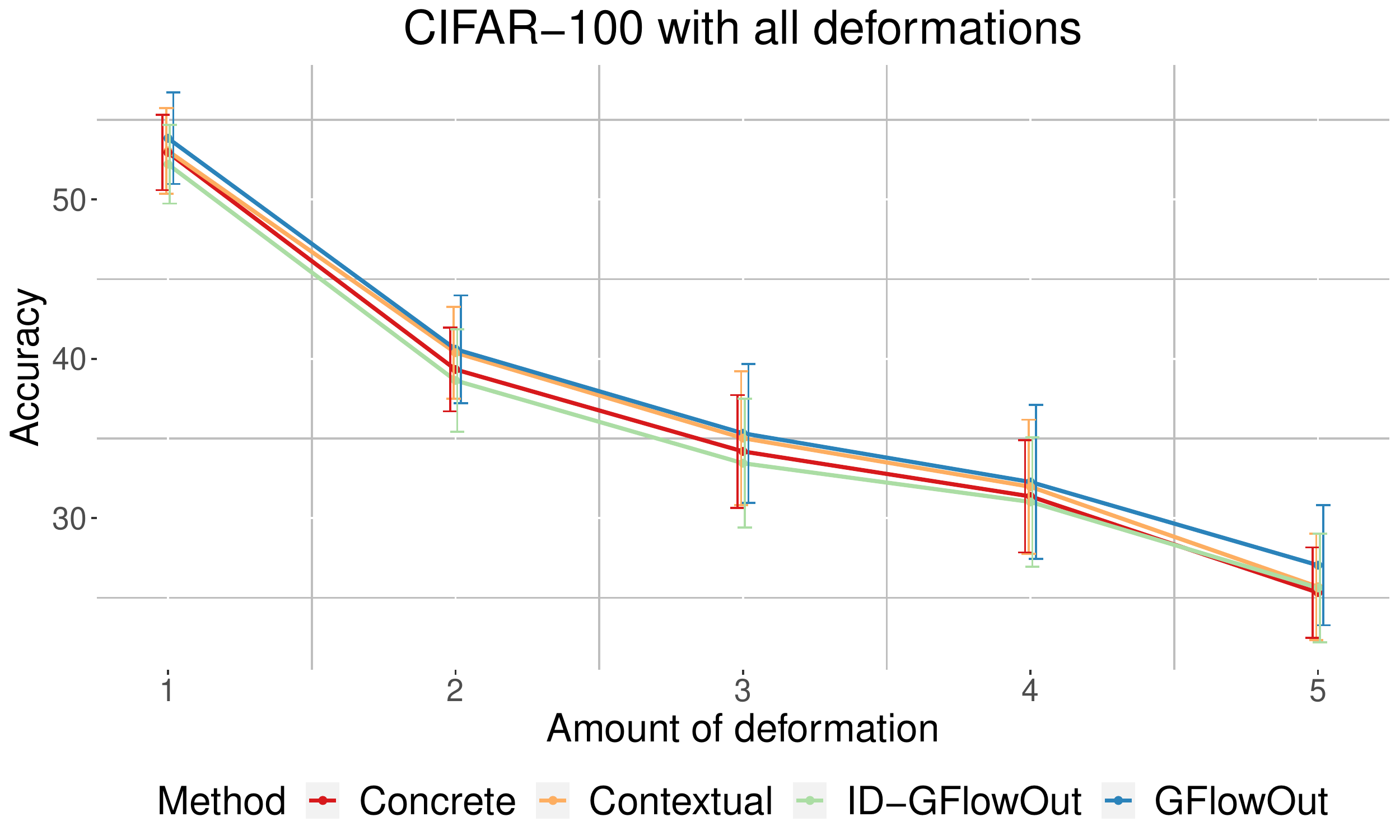}
    \includegraphics[width=0.38\linewidth]{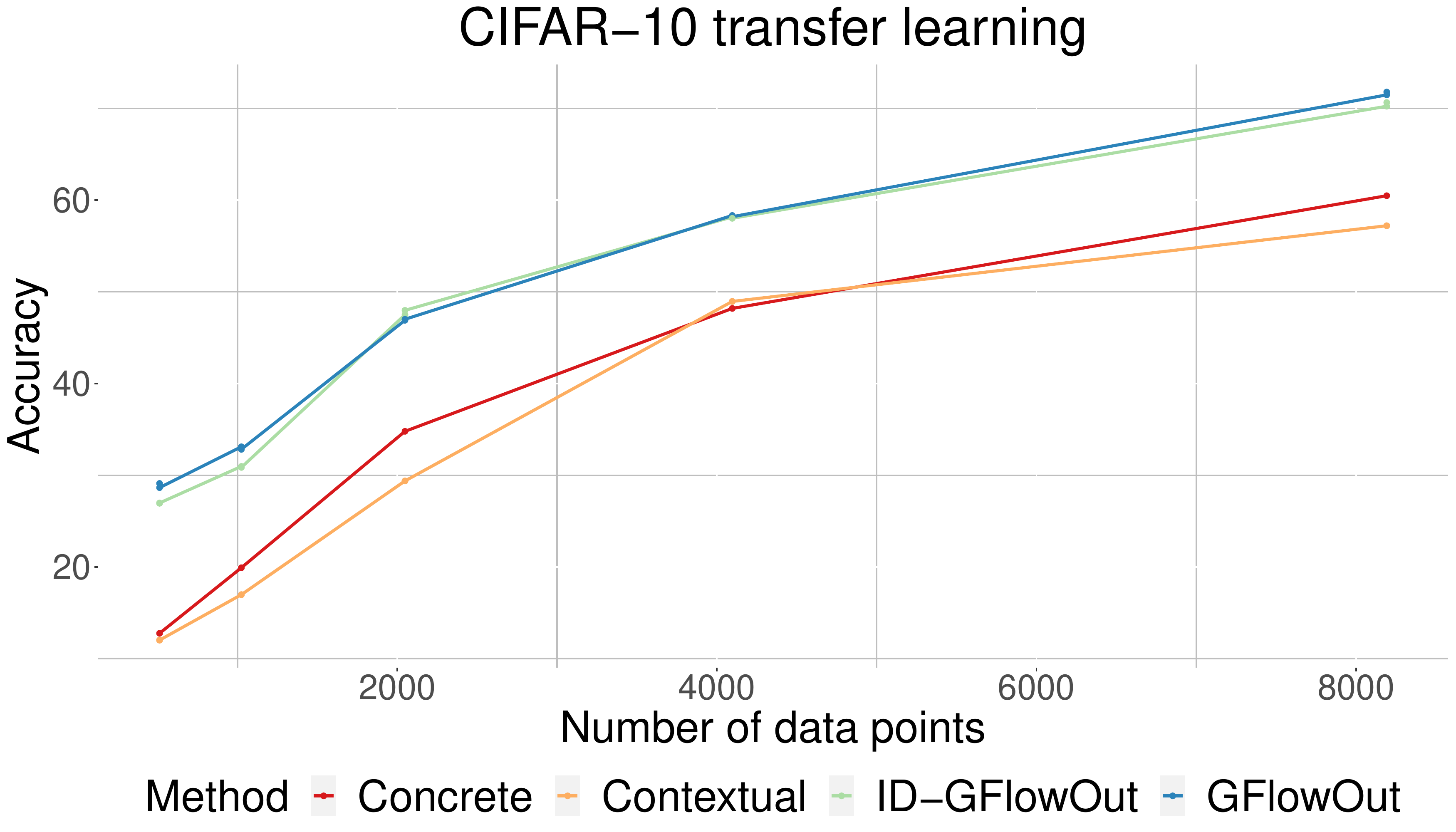}
    \includegraphics[width=0.38\linewidth]{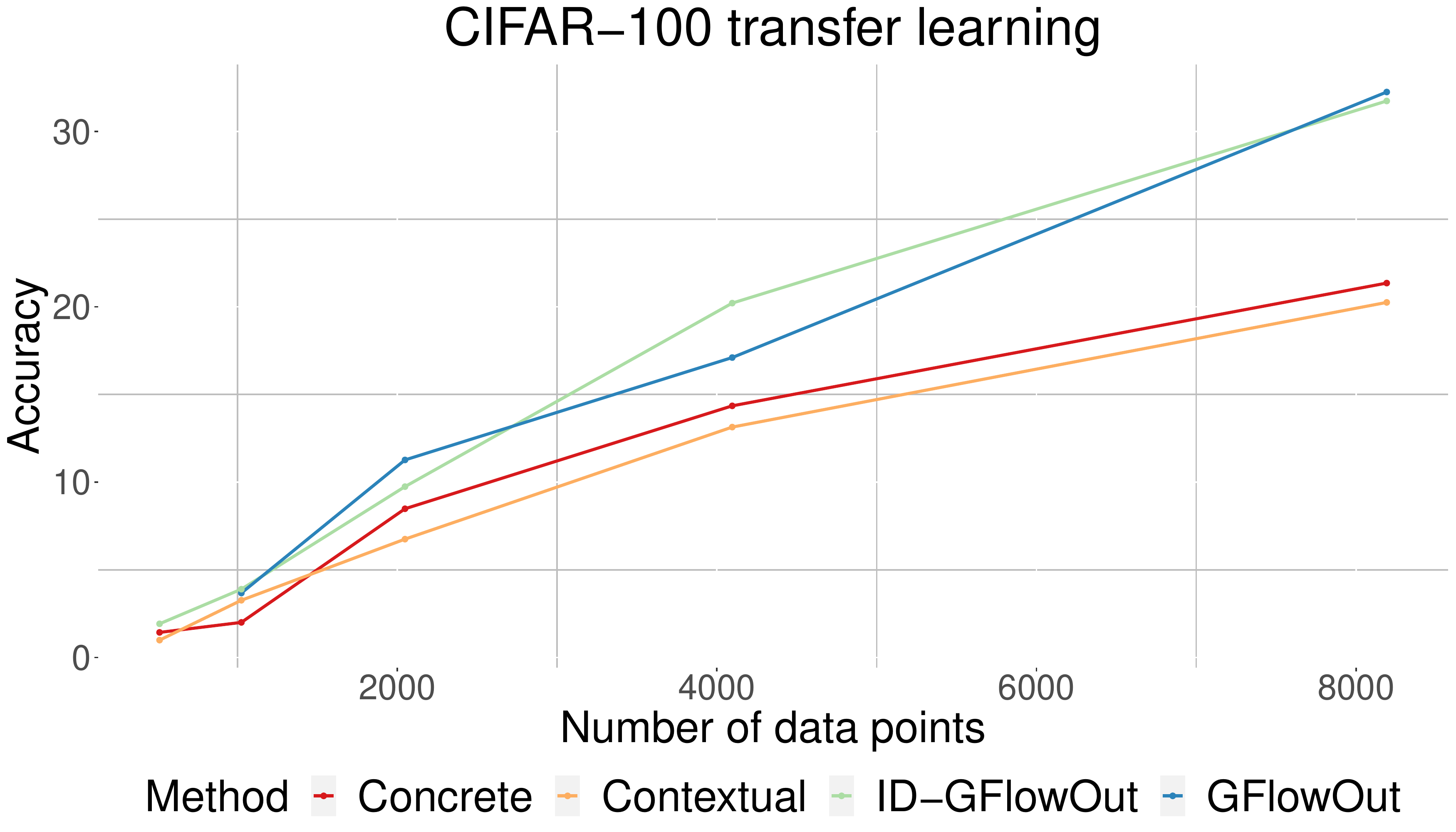}

    \caption{\emph{Top}: Evaluating the robustness of ResNet-18 models, trained with different dropout methods, to different amounts of deformation including SNOW deformation, FROST deformation or Gaussian noise on CIFAR-10/CIFAR-100 at test time. See Figure \ref{augmentation_CIFAR-10} and \ref{augmentation_CIFAR-100} in Appendix for detailed results. \emph{Bottom}: Evaluating the transfer learning performance of ResNet-18 models trained on CIFAR-10/CIFAR-100 with label noise and fine-tuned on varying amounts of clean data (i.e., without any label noise). 
    }
    \label{fig:NoiseNtransferlearning}
    \vspace{-4mm}
\end{figure*}

\textbf{Robustness to data distribution shift.} To evaluate the robustness of GFlowOut to distribution shifts between the train and test data, we study its predictive performance on OOD examples. We conduct experiments on MNIST, CIFAR-10, and CIFAR-100 datasets with different types and levels of deformations. For MNIST, we train a two-layer MLP with 300 and 100 units respectively and evaluate predictions on MNIST images rotated by a uniformly sampled angle ($0-360^{\circ}$). Similarly, we use the ResNet-18~\cite{he2016deep} models for the CIFAR-10/CIFAR-100 datasets and evaluate their robustness to distribution shifts induced by random rotations. Additionally, we consider ``Snow", ``Frost" and Gaussian noises image corruptions~\cite{hendrycks2019benchmarking}, and analyze the robustness of models to each type of deformation applied with varying intensities. We consider both GFlowOut and ID-GFlowOut variants and as baselines use Random Dropout (Standard Bernoulli Dropout)~\cite{hinton2012improving}, Contextual Dropout~\cite{fan2021contextual} and Concrete Dropout~\cite{gal2017concrete}. The results, as summarized in Table~\ref{tab:robustness}, show that models trained using GFlowOut are in general more robust to random rotations, and GFlowOut outperforms (or at least matches the performance of) baselines in five out of six experiments with different levels of corruption (Figure~\ref{fig:NoiseNtransferlearning}, Appendix Figure~\ref{augmentation_CIFAR-10} and Figure~\ref{augmentation_CIFAR-100}). These observations suggest that models trained with GFlowOut are more robust to distribution shifts as compared to the baselines. Better generalization performance to distribution shifts indicates that GFlowOut potentially learns a better approximation of the Bayesian posterior over the model parameters.

\begin{table}[]
\caption{Performance on clean and corrupted data to evaluate the robustness of models trained with different dropout methods to random image rotations at test time.}
\centering
\begin{tabular}{llll}
\toprule
\textbf{Data}     & \textbf{Method}       & \textbf{Acc.} & \textbf{Acc.(rotated)} \\ \midrule
\multirow{5}{*}{CIFAR-10}           & Concrete              & 90.13                  & 30.68                    \\ 
           & Contextual            & 90.12                  & 29.11                    \\ 
           & Random                & 91.47                  & 27.04                    \\ 
           & ID-GFlowOut          & \textbf{91.85}                  & 30.57                    \\ 
  & GFlowOut     & 91.52         & \textbf{31.00}            \\ \midrule 
\multirow{5}{*}{CIFAR-100}          & Concrete              & 62.49                  & 16.43                    \\ 
          & Contextual            & 62.30                   & 15.93                    \\ 
          & Random                & 67.19                  & 15.70                     \\ 
          & ID-GFlowOut          & \textbf{69.99}                  & 16.29                    \\ 
 & GFlowOut     & 69.80         & \textbf{17.01}           \\ \midrule 
\multirow{5}{*}{MNIST}             & Concrete              & 97.36                  & 66.10                    \\ 
             & Contextual            & \textbf{98.15}                  & 66.20                     \\ 
             & Random                & 87.38                  & 43.96                    \\ 
    & ID-GFlowOut & 97.05          & \textbf{70.19}           \\ 
             & GFlowOut              & 96.75                  & 66.41                    \\ \bottomrule
\end{tabular}
\label{tab:robustness}
\end{table}

\textbf{Visual Question Answering task using transformer architecture}
To evaluate GFlowOut on large-scale tasks with larger models, we consider a transformer-based multi-modal architecture MCAN~\citep{yu2019deep} for the Visual Question Answering (VQA) task, following \citet{fan2021contextual}. The task involves answering a textual question related to the content of a given image. There are three types of questions in the task, namely binary yes/no questions, numerical questions, and other questions. Dropout is applied on cross-modal attention between images and texts, within data type self-attention, and the feed-forward layers after attention. Our experimental results in Table~\ref{tab:VQA} suggest that GFlowOut either outperforms or matches the performance of contextual and concrete dropout when tested on generalization to noisy dataset where a Gaussian noise is added to the visual inputs \citep{fan2021contextual}.

\begin{table*}
\vspace{-1mm}
\centering
\caption{Performance on different question types in Visual Question Answering task with a Transformer-based model trained with different methods.}
\resizebox{0.88\textwidth}{!}{
\begin{tabular}{llcccc}
\toprule
\textbf{Method} & \textbf{Test Set} & \textbf{Acc.(All)} & \textbf{Acc.(Yes/No)} & \textbf{Acc.(Number)} & \textbf{Acc.(Other)} \\ \midrule
Ide-GFlowOut       &           &  66.66     & 84.21        & 48.99        & \textbf{58.42}       \\ 
GFlowOut        & \multirow{3}{*}{Original}           & 66.12     & 83.91        & 49.01        & 58.33       \\ 
Contextual      &           & 66.89              & 84.48                 & \textbf{49.04}                 & 58.24                \\ 
Concrete        &            & \textbf{66.92}              & \textbf{84.51}                 & 48.66                 & 58.38                \\ \midrule
Ide-GFlowOut        &           & 50.27     & \textbf{74.01}        & 32.16        & 36.12                \\ 
GFlowOut        & \multirow{3}{*}{Noisy}              & \textbf{50.33}     & 73.31        & \textbf{32.64}        & \textbf{40.17}                \\ 
Contextual      &              & 49.72              & 73.81                 & 32.4                  & 35.97                \\ 
Concrete        &              & 50.2               & 73.5                  & 31.45                 & 37.39       \\ \bottomrule
\end{tabular}
}
\label{tab:VQA}
\vspace{-0mm}
\end{table*} 

\textbf{Uncertainty estimation for out-of-distribution detection.} Another way to evaluate the quality of the learned posterior is to analyze the uncertainty estimates on a downstream task. We consider the standard task of using uncertainty estimates for detecting out-of-distribution (OOD) examples~\cite{nado2021uncertainty}. The intuition is that a well-calibrated model should produce uncertainty in predictions on OOD examples. This can be useful in cases where difficult OOD examples can be delegated to humans for more careful consideration. As in the previous experiments, we consider ResNet-18 models for CIFAR-10/CIFAR-100 classification and compute uncertainty estimates on the CIFAR-10/CIFAR-100 and SVHN (OOD) test sets. Uncertainty for prediction on each example is calculated using the Dempster-Shafer metric~\cite{sensoy2018evidential}. For baselines, we consider Contextual Dropout and Concrete Dropout, along with standard MC Dropout and Deep Ensembles which are strong baselines for this task. We run the experiment with 5 seeds and report the mean and standard error. 
We study both GFlowOut with sample-dependent information and ID-GFlowOut with only sample-independent information. In Table~\ref{tab:uncertainty}, we present AUPR and AUROC for in-distribution classification (CIFAR-10 and CIFAR-100) and OOD classification (SVHN) using the uncertainty estimates from each method. We observe that GFlowOut outperforms the other dropout baselines with both CIFAR-10 and CIFAR-100 as the training dataset, indicating that sample-dependent information used in GFlowOut results in more calibrated uncertainty estimates. ID-GFlowOut performs well on CIFAR-100 but performs poorly on CIFAR-10. Results of deep ensembles, which is a widely used state-of-the-art uncertainty estimation method, is also reported in Table~\ref{tab:uncertainty} for comparison.

\textbf{Adaptation after training on noisy data.} Next, we evaluate the ability of models trained with GFlowOut to adapt quickly after being trained on noisy data. Concretely, during training we add label noise i.e., we randomly assign the labels for a fraction (30\%) of points in the training set and then re-train the classifier on a small fraction of the dataset with clean labels. We adopt the same experimental setup as the previous experiments. We consider ResNet-18 models trained on CIFAR-10/CIFAR-100 datasets. As baselines, we again use Contextual Dropout and Concrete Dropout. Figure~\ref{fig:NoiseNtransferlearning} shows that the models trained with  GFlowOut perform adapt faster than the dropout methods we use as a baseline. We also observe that both the sample-dependent and sample-independent variants of GFlowOut achieve similar performance.

\begin{table}[ht]
\vspace{-2mm}
\centering
\caption{Performance on the out-of-distribution (OOD) detection task indicates the posterior approximated by GFlowOut provides better uncertainty estimates.
}
\resizebox{0.48\textwidth}{!}{
\begin{tabular}{llll}
\toprule
\textbf{Data}      & \textbf{Method}       & \textbf{AUROC} & \textbf{AUPR}  \\ \midrule
\multirow{6}{*}{CIFAR-10}  & Concrete     & 0.909$\pm$ 0.021 & 0.872$\pm$ 0.017 \\
          & Contextual   & 0.915$\pm$ 0.011 & 0.874$\pm$ 0.014 \\
          & MC Dropout    & 0.882$\pm$ 0.009 & 0.869$\pm$ 0.010 \\
          & Deep Ensembles & 0.935$\pm$ 0.007 & 0.912$\pm$ 0.006 \\
          & ID-GFlowOut & 0.781$\pm$ 0.021 & 0.843$\pm$ 0.018 \\
          & GFlowOut     & \textbf{0.955$\pm$ 0.013} & \textbf{0.924$\pm$ 0.019} \\ \midrule
\multirow{6}{*}{CIFAR-100} & Concrete     & 0.795$\pm$ 0.017 & 0.691$\pm$ 0.021 \\
          & Contextual   & 0.812$\pm$ 0.013 & 0.728$\pm$ 0.014\\
          & MC Dropout    & 0.783$\pm$ 0.011 & 0.715$\pm$ 0.021 \\
          & Deep Ensembles & \textbf{0.842$\pm$ 0.008} & 0.731$\pm$ 0.017 \\
          & ID-GFlowOut & 0.819$\pm$ 0.008 & 0.707$\pm$ 0.012 \\
          & GFlowOut     & \textbf{0.839$\pm$ 0.010} & \textbf{0.741$\pm$ 0.012} \\ \bottomrule
\end{tabular}
}
\label{tab:uncertainty}
\vspace{-4mm}
\end{table}

\textbf{Application on real-world clinical data.} We explore the competence of GFlowOut as a probabilistic tool for solving real-world problems. In intensive care units (ICUs), the ability to forecast the mortality of patients can help clinicians to allocate limited resources to help the individuals at the highest risk. However, to respect patient privacy, most hospitals have access only to data for a limited number of patients that is anonymized and available for training predictive models. Moreover, there are stringent regulations on the exchange of medical records among hospitals. To enable data-driven decision-making in these critical scenarios, we consider learning probabilistic classifiers for the problem of mortality predictions. We use patients' medication usage in the first 48 hrs of ICU stay to make a binary prediction of mortality during the stay. We emphasize that the predictions are meant to help decision-makers (doctors) in making clinical decisions rather than being used directly. We use a 3-layer MLP trained with 4500 patients' ICU records, including 158 deaths, from one hospital, and tested with data from another hospital (4018 patients, 147 deaths). Records of both hospitals are obtained from the eICU database~\cite{pollard2018eicu}. This task encompasses several important challenges in applying machine learning tools to real-world tasks: (1) the underlying prediction task is extremely difficult due to limited information and complex case-specific clinical details, (2) the data distribution is severely imbalanced as deaths are rare events, (3) limited training data resulting in a complex posterior, and (4) large distribution shifts between hospitals. Overall, this cross-hospital task setup is quite challenging. 
Our results in Table~\ref{tab:clinical}, show that GFlowOut significantly outperforms the baselines, in all the metrics. While the margins may appear to be small, in the context of real-world decision-making, they can have a significant impact. The findings demonstrate GFlowOut's effectiveness in addressing risk-averse real-world problems.

\begin{table}[]
\vspace{-2mm}
\centering
\caption{Performance of methods on cross-hospital mortality prediction on real-world clinical data from ICUs demonstrates superior performance of GFlowOut. 
}
\begin{tabular}{llll}
\toprule
\textbf{Method} & \textbf{F1 (Macro)} & \textbf{Precision} & \textbf{Recall} \\ \midrule
Concrete        & 0.528               & 0.524              & 0.641           \\ 
Contextual      & 0.521               & 0.518              & 0.659           \\ 
Random          & 0.49                & 0.5                & 0.499           \\ 
ID-GFlowOut         & 0.499               & 0.506              & 0.534           \\ 
GFlowOut        & \textbf{0.536}      & \textbf{0.528}     & \textbf{0.681}  \\ \bottomrule
\end{tabular}
\label{tab:clinical}
\vspace{-6mm}
\end{table}




\section{Conclusion}
In this work, we propose GFlowOut, to learn the posterior distribution over dropout masks in a neural network. We evaluate GFlowOut on various downstream tasks such as uncertainty estimation, robustness to distribution shift, and transfer learning, using both benchmark datasets and real-world clinical datasets. Our empirical results show the favorable performance of GFlowOut over related methods like Concrete and Contextual Dropout. Future work should involve combining top-down and bottom-up dropout strategies, applying GFlowOut on larger models with complex architectures, and using it to promote exploration in RL problems where accurate estimation of posterior has shown to enhance sample efficiency \cite{osband2013more}.

\section*{Author contributions}
D.L., Y.B. and K.K. initialized the project. D.L., M.J., B.D., C.E., Q.S. and S.L. contributed to implementation and experiments of the project. D.L., M.J. and A.G. designed the experimental studies.  D.L., Y.B., A.G., N.M., M.J., X.J. and S.L. contributed to conceptualization of the project. N.M., S.L., K.K., D.Z., D.L., M.J., Q.S., N.H. and Y.B. contributed to the mathematical parts of the project. D.L., A.G. and M.J. coordinated the project. Y.B. supervised the whole project and designed the whole framework. All authors contributed to the writing of the manuscript.

\section*{Acknowledgments}
The authors thank CIFAR, Samsung and IVADO for funding and NVIDIA for equipment.


\bibliography{Reference}

\begin{thebibliography}{48}
\providecommand{\natexlab}[1]{#1}
\providecommand{\url}[1]{\texttt{#1}}
\expandafter\ifx\csname urlstyle\endcsname\relax
  \providecommand{\doi}[1]{doi: #1}\else
  \providecommand{\doi}{doi: \begingroup \urlstyle{rm}\Url}\fi

\bibitem[Ba \& Frey(2013)Ba and Frey]{ba2013adaptive}
Ba, J. and Frey, B.
\newblock Adaptive dropout for training deep neural networks.
\newblock \emph{Neural Information Processing Systems (NIPS)}, 2013.

\bibitem[Bengio et~al.(2021{\natexlab{a}})Bengio, Jain, Korablyov, Precup, and
  Bengio]{bengio2021flow}
Bengio, E., Jain, M., Korablyov, M., Precup, D., and Bengio, Y.
\newblock Flow network based generative models for non-iterative diverse
  candidate generation.
\newblock \emph{Neural Information Processing Systems (NeurIPS)},
  2021{\natexlab{a}}.

\bibitem[Bengio et~al.(2021{\natexlab{b}})Bengio, Deleu, Hu, Lahlou, Tiwari,
  and Bengio]{bengio2021gflownet}
Bengio, Y., Deleu, T., Hu, E.~J., Lahlou, S., Tiwari, M., and Bengio, E.
\newblock Gflownet foundations.
\newblock \emph{arXiv preprint arXiv:2111.09266}, 2021{\natexlab{b}}.

\bibitem[Bhatt et~al.(2021)Bhatt, Antor{\'a}n, Zhang, Liao, Sattigeri,
  Fogliato, Melan{\c{c}}on, Krishnan, Stanley, Tickoo,
  et~al.]{bhatt2021uncertainty}
Bhatt, U., Antor{\'a}n, J., Zhang, Y., Liao, Q.~V., Sattigeri, P., Fogliato,
  R., Melan{\c{c}}on, G., Krishnan, R., Stanley, J., Tickoo, O., et~al.
\newblock Uncertainty as a form of transparency: Measuring, communicating, and
  using uncertainty.
\newblock In \emph{Proceedings of the 2021 AAAI/ACM Conference on AI, Ethics,
  and Society}, pp.\  401--413, 2021.

\bibitem[Boluki et~al.(2020)Boluki, Ardywibowo, Dadaneh, Zhou, and
  Qian]{boluki2020learnable}
Boluki, S., Ardywibowo, R., Dadaneh, S.~Z., Zhou, M., and Qian, X.
\newblock Learnable {Bernoulli} dropout for {Bayesian} deep learning.
\newblock \emph{Artificial Intelligence and Statistics (AISTATS)}, 2020.

\bibitem[Damianou \& Lawrence(2013)Damianou and Lawrence]{damianou2013deep}
Damianou, A. and Lawrence, N.~D.
\newblock Deep {Gaussian} processes.
\newblock \emph{Artificial Intelligence and Statistics (AISTATS)}, 2013.

\bibitem[Daxberger et~al.(2021)Daxberger, Nalisnick, Allingham, Antor{\'a}n,
  and Hern{\'a}ndez-Lobato]{daxberger2021bayesian}
Daxberger, E., Nalisnick, E., Allingham, J.~U., Antor{\'a}n, J., and
  Hern{\'a}ndez-Lobato, J.~M.
\newblock Bayesian deep learning via subnetwork inference.
\newblock \emph{International Conference on Machine Learning (ICML)}, 2021.

\bibitem[Deleu et~al.(2022)Deleu, G{\'o}is, Emezue, Rankawat, Lacoste-Julien,
  Bauer, and Bengio]{deleu2022bayesian}
Deleu, T., G{\'o}is, A., Emezue, C., Rankawat, M., Lacoste-Julien, S., Bauer,
  S., and Bengio, Y.
\newblock Bayesian structure learning with generative flow networks.
\newblock \emph{Uncertainty in Artificial Intelligence (UAI)}, 2022.

\bibitem[Fan et~al.(2021)Fan, Zhang, Tanwisuth, Qian, and
  Zhou]{fan2021contextual}
Fan, X., Zhang, S., Tanwisuth, K., Qian, X., and Zhou, M.
\newblock Contextual dropout: An efficient sample-dependent dropout module.
\newblock \emph{International Conference on Learning Representations (ICLR)},
  2021.

\bibitem[Foong et~al.(2020)Foong, Burt, Li, and
  Turner]{foong2020expressiveness}
Foong, A., Burt, D., Li, Y., and Turner, R.
\newblock On the expressiveness of approximate inference in {Bayesian} neural
  networks.
\newblock \emph{Neural Information Processing Systems (NeurIPS)}, 2020.

\bibitem[Fort et~al.(2019)Fort, Hu, and Lakshminarayanan]{fort2019deep}
Fort, S., Hu, H., and Lakshminarayanan, B.
\newblock Deep ensembles: A loss landscape perspective.
\newblock \emph{arXiv preprint arXiv:1912.02757}, 2019.

\bibitem[Gal \& Ghahramani(2016)Gal and Ghahramani]{gal2016dropout}
Gal, Y. and Ghahramani, Z.
\newblock Dropout as a {Bayesian} approximation: Representing model uncertainty
  in deep learning.
\newblock \emph{International Conference on Machine Learning (ICML)}, 2016.

\bibitem[Gal et~al.(2017)Gal, Hron, and Kendall]{gal2017concrete}
Gal, Y., Hron, J., and Kendall, A.
\newblock Concrete dropout.
\newblock \emph{Neural Information Processing Systems (NIPS)}, 30, 2017.

\bibitem[Ghiasi et~al.(2018)Ghiasi, Lin, and Le]{ghiasi2018dropblock}
Ghiasi, G., Lin, T.-Y., and Le, Q.~V.
\newblock Dropblock: A regularization method for convolutional networks.
\newblock \emph{Neural Information Processing Systems (NeurIPS)}, 2018.

\bibitem[Guo et~al.(2017)Guo, Pleiss, Sun, and Weinberger]{guo2017calibration}
Guo, C., Pleiss, G., Sun, Y., and Weinberger, K.~Q.
\newblock On calibration of modern neural networks.
\newblock \emph{International Conference on Machine Learning (ICML)}, 2017.

\bibitem[He et~al.(2016)He, Zhang, Ren, and Sun]{he2016deep}
He, K., Zhang, X., Ren, S., and Sun, J.
\newblock Deep residual learning for image recognition.
\newblock \emph{Computer Vision and Pattern Recognition (CVPR)}, 2016.

\bibitem[Hendrycks \& Dietterich(2019)Hendrycks and
  Dietterich]{hendrycks2019benchmarking}
Hendrycks, D. and Dietterich, T.
\newblock Benchmarking neural network robustness to common corruptions and
  perturbations.
\newblock \emph{International Conference on Learning Representations (ICLR)},
  2019.

\bibitem[Hinton et~al.(2012)Hinton, Srivastava, Krizhevsky, Sutskever, and
  Salakhutdinov]{hinton2012improving}
Hinton, G.~E., Srivastava, N., Krizhevsky, A., Sutskever, I., and
  Salakhutdinov, R.~R.
\newblock Improving neural networks by preventing co-adaptation of feature
  detectors.
\newblock \emph{arXiv preprint arXiv:1207.0580}, 2012.

\bibitem[Jain et~al.(2022)Jain, Bengio, Hernandez-Garcia, Rector-Brooks,
  Dossou, Ekbote, Fu, Zhang, Kilgour, Zhang, Simine, Das, and
  Bengio]{jain2022biological}
Jain, M., Bengio, E., Hernandez-Garcia, A., Rector-Brooks, J., Dossou, B.~F.,
  Ekbote, C.~A., Fu, J., Zhang, T., Kilgour, M., Zhang, D., Simine, L., Das,
  P., and Bengio, Y.
\newblock Biological sequence design with gflownets.
\newblock \emph{International Conference on Machine Learning (ICML)}, 2022.

\bibitem[Jain et~al.(2023)Jain, Lahlou, Nekoei, Butoi, Bertin, Rector-Brooks,
  Korablyov, and Bengio]{jain2021deup}
Jain, M., Lahlou, S., Nekoei, H., Butoi, V., Bertin, P., Rector-Brooks, J.,
  Korablyov, M., and Bengio, Y.
\newblock {DEUP}: Direct epistemic uncertainty prediction.
\newblock \emph{Transactions on Machine Learning Research (TMLR)}, 2023.

\bibitem[Kingma et~al.(2015)Kingma, Salimans, and
  Welling]{kingma2015variational}
Kingma, D.~P., Salimans, T., and Welling, M.
\newblock Variational dropout and the local reparameterization trick.
\newblock \emph{Neural Information Processing Systems (NIPS)}, 2015.

\bibitem[Kuleshov et~al.(2018)Kuleshov, Fenner, and
  Ermon]{kuleshov2018accurate}
Kuleshov, V., Fenner, N., and Ermon, S.
\newblock Accurate uncertainties for deep learning using calibrated regression.
\newblock \emph{International Conference on Machine Learning (ICML)}, 2018.

\bibitem[Le~Folgoc et~al.(2021)Le~Folgoc, Baltatzis, Desai, Devaraj, Ellis,
  Manzanera, Nair, Qiu, Schnabel, and Glocker]{le2021mc}
Le~Folgoc, L., Baltatzis, V., Desai, S., Devaraj, A., Ellis, S., Manzanera, O.
  E.~M., Nair, A., Qiu, H., Schnabel, J., and Glocker, B.
\newblock Is {MC} dropout bayesian?
\newblock \emph{arXiv preprint arXiv:2110.04286}, 2021.

\bibitem[Lee et~al.(2020)Lee, Nam, Yang, and Hwang]{lee2020meta}
Lee, H.~B., Nam, T., Yang, E., and Hwang, S.~J.
\newblock Meta dropout: Learning to perturb latent features for generalization.
\newblock \emph{International Conference on Learning Representations (ICLR)},
  2020.

\bibitem[Lotfi et~al.(2022)Lotfi, Izmailov, Benton, Goldblum, and
  Wilson]{lotfi2022bayesian}
Lotfi, S., Izmailov, P., Benton, G., Goldblum, M., and Wilson, A.~G.
\newblock Bayesian model selection, the marginal likelihood, and
  generalization.
\newblock \emph{International Conference on Machine Learning (ICML)}, 2022.

\bibitem[MacKay(1992)]{mackay1992practical}
MacKay, D.~J.
\newblock A practical {Bayesian} framework for backpropagation networks.
\newblock \emph{Neural Computation}, 4\penalty0 (3):\penalty0 448--472, 1992.

\bibitem[Madan et~al.(2023)Madan, Rector-Brooks, Korablyov, Bengio, Jain, Nica,
  Bosc, Bengio, and Malkin]{madan2022learning}
Madan, K., Rector-Brooks, J., Korablyov, M., Bengio, E., Jain, M., Nica, A.,
  Bosc, T., Bengio, Y., and Malkin, N.
\newblock Learning {GFlowNets} from partial episodes for improved convergence
  and stability.
\newblock \emph{International Conference on Machine Learning (ICML)}, 2023.

\bibitem[Malkin et~al.(2022)Malkin, Jain, Bengio, Sun, and
  Bengio]{malkin2022trajectory}
Malkin, N., Jain, M., Bengio, E., Sun, C., and Bengio, Y.
\newblock Trajectory balance: Improved credit assignment in {GFlowNets}.
\newblock \emph{Neural Information Processing Systems (NeurIPS)}, 2022.

\bibitem[Malkin et~al.(2023)Malkin, Lahlou, Deleu, Ji, Hu, Everett, Zhang, and
  Bengio]{malkin2022gflownets}
Malkin, N., Lahlou, S., Deleu, T., Ji, X., Hu, E., Everett, K., Zhang, D., and
  Bengio, Y.
\newblock {GFlowNets} and variational inference.
\newblock \emph{International Conference on Learning Representations (ICLR)},
  2023.

\bibitem[Nado et~al.(2021)Nado, Band, Collier, Djolonga, Dusenberry, Farquhar,
  Filos, Havasi, Jenatton, Jerfel, Liu, Mariet, Nixon, Padhy, Ren, Rudner, Wen,
  Wenzel, Murphy, Sculley, Lakshminarayanan, Snoek, Gal, and
  Tran]{nado2021uncertainty}
Nado, Z., Band, N., Collier, M., Djolonga, J., Dusenberry, M., Farquhar, S.,
  Filos, A., Havasi, M., Jenatton, R., Jerfel, G., Liu, J., Mariet, Z., Nixon,
  J., Padhy, S., Ren, J., Rudner, T., Wen, Y., Wenzel, F., Murphy, K., Sculley,
  D., Lakshminarayanan, B., Snoek, J., Gal, Y., and Tran, D.
\newblock {Uncertainty Baselines}: Benchmarks for uncertainty \& robustness in
  deep learning.
\newblock \emph{arXiv preprint arXiv:2106.04015}, 2021.

\bibitem[Neal(2012)]{neal2012bayesian}
Neal, R.~M.
\newblock \emph{Bayesian learning for neural networks}, volume 118.
\newblock Springer Science \& Business Media, 2012.

\bibitem[Nguyen et~al.(2015)Nguyen, Yosinski, and Clune]{nguyen2015deep}
Nguyen, A., Yosinski, J., and Clune, J.
\newblock Deep neural networks are easily fooled: High confidence predictions
  for unrecognizable images.
\newblock \emph{Computer Vision and Pattern Recognition (CVPR)}, 2015.

\bibitem[Nguyen et~al.(2021)Nguyen, Nguyen, Nguyen, Than, Bui, and
  Ho]{nguyen2021structured}
Nguyen, S., Nguyen, D., Nguyen, K., Than, K., Bui, H., and Ho, N.
\newblock Structured dropout variational inference for {Bayesian} neural
  networks.
\newblock \emph{Neural Information Processing Systems (NeurIPS)}, 2021.

\bibitem[Nica et~al.(2022)Nica, Jain, Bengio, Liu, Korablyov, Bronstein, and
  Bengio]{nica2022evaluating}
Nica, A.~C., Jain, M., Bengio, E., Liu, C.-H., Korablyov, M., Bronstein, M.~M.,
  and Bengio, Y.
\newblock Evaluating generalization in gflownets for molecule design.
\newblock \emph{ICLR 2022 Machine Learning for Drug Discovery workshop}, 2022.

\bibitem[Osband et~al.(2013)Osband, Russo, and Van~Roy]{osband2013more}
Osband, I., Russo, D., and Van~Roy, B.
\newblock (more) efficient reinforcement learning via posterior sampling.
\newblock \emph{Neural Information Processing Systems (NIPS)}, 2013.

\bibitem[Ovadia et~al.(2019)Ovadia, Fertig, Ren, Nado, Sculley, Nowozin,
  Dillon, Lakshminarayanan, and Snoek]{ovadia2019can}
Ovadia, Y., Fertig, E., Ren, J., Nado, Z., Sculley, D., Nowozin, S., Dillon,
  J., Lakshminarayanan, B., and Snoek, J.
\newblock Can you trust your model's uncertainty? evaluating predictive
  uncertainty under dataset shift.
\newblock \emph{Neural Information Processing Systems (NeurIPS)}, 2019.

\bibitem[Pan et~al.(2023)Pan, Zhang, Courville, Huang, and
  Bengio]{Pan2022GenerativeAF}
Pan, L., Zhang, D., Courville, A.~C., Huang, L., and Bengio, Y.
\newblock Generative augmented flow networks.
\newblock \emph{International Conference on Learning Representations (ICLR)},
  2023.

\bibitem[Park \& Kwak(2016)Park and Kwak]{park2016analysis}
Park, S. and Kwak, N.
\newblock Analysis on the dropout effect in convolutional neural networks.
\newblock \emph{Asian Conference on Computer Vision}, 2016.

\bibitem[Pham \& Le(2021)Pham and Le]{pham2021autodropout}
Pham, H. and Le, Q.
\newblock Autodropout: Learning dropout patterns to regularize deep networks.
\newblock \emph{Association for the Advancement of Artificial Intelligence
  (AAAI)}, 2021.

\bibitem[Pollard et~al.(2018)Pollard, Johnson, Raffa, Celi, Mark, and
  Badawi]{pollard2018eicu}
Pollard, T.~J., Johnson, A.~E., Raffa, J.~D., Celi, L.~A., Mark, R.~G., and
  Badawi, O.
\newblock The {eICU Collaborative Research Database}, a freely available
  multi-center database for critical care research.
\newblock \emph{Scientific data}, 5\penalty0 (1):\penalty0 1--13, 2018.

\bibitem[Sensoy et~al.(2018)Sensoy, Kaplan, and Kandemir]{sensoy2018evidential}
Sensoy, M., Kaplan, L., and Kandemir, M.
\newblock Evidential deep learning to quantify classification uncertainty.
\newblock \emph{Neural Information Processing Systems (NeurIPS)}, 2018.

\bibitem[Sutton \& Barto(2018)Sutton and Barto]{sutton2018reinforcement}
Sutton, R.~S. and Barto, A.~G.
\newblock \emph{Reinforcement learning: An introduction}.
\newblock MIT press, 2018.

\bibitem[Wilson \& Izmailov(2020)Wilson and Izmailov]{wilson2020bayesian}
Wilson, A.~G. and Izmailov, P.
\newblock Bayesian deep learning and a probabilistic perspective of
  generalization.
\newblock \emph{Neural Information Processing Systems (NeurIPS)}, 2020.

\bibitem[Xie et~al.(2019)Xie, Ma, Zhang, Xue, Tan, and Guo]{xie2019soft}
Xie, J., Ma, Z., Zhang, G., Xue, J.-H., Tan, Z.-H., and Guo, J.
\newblock Soft dropout and its variational bayes approximation.
\newblock \emph{Machine Learning for Signal Processing (MLSP)}, 2019.

\bibitem[Yang et~al.(2020)Yang, Tang, Torun, Becker, Hejase, and
  Swaminathan]{yang2020rx}
Yang, X., Tang, J., Torun, H.~M., Becker, W.~D., Hejase, J.~A., and
  Swaminathan, M.
\newblock Rx equalization for a high-speed channel based on bayesian active
  learning using dropout.
\newblock \emph{Electrical Performance of Electronic Packaging and Systems
  (EPEPS)}, 2020.

\bibitem[Yu et~al.(2019)Yu, Yu, Cui, Tao, and Tian]{yu2019deep}
Yu, Z., Yu, J., Cui, Y., Tao, D., and Tian, Q.
\newblock Deep modular co-attention networks for visual question answering.
\newblock \emph{Computer Vision and Pattern Recognition (CVPR)}, 2019.

\bibitem[Zhang et~al.(2022{\natexlab{a}})Zhang, Chen, Malkin, and
  Bengio]{zhang2022unifying}
Zhang, D., Chen, R.~T., Malkin, N., and Bengio, Y.
\newblock Unifying generative models with gflownets.
\newblock \emph{arXiv preprint arXiv:2209.02606}, 2022{\natexlab{a}}.

\bibitem[Zhang et~al.(2022{\natexlab{b}})Zhang, Malkin, Liu, Volokhova,
  Courville, and Bengio]{zhang2022generative}
Zhang, D., Malkin, N., Liu, Z., Volokhova, A., Courville, A., and Bengio, Y.
\newblock Generative flow networks for discrete probabilistic modeling.
\newblock \emph{International Conference on Machine Learning (ICML)},
  2022{\natexlab{b}}.

\end{thebibliography}
\bibliographystyle{icml2023}

\newpage
\appendix
\onecolumn

\section{Appendix}

\subsubsection{Hyperparameters}
Hyperparameters of the ``backbone" ResNet and Transformer models were obtained from published baselines or architectures~\citep{he2016deep,yu2019deep,fan2021contextual,gal2016dropout,gal2017concrete}.  Several GFlowNet-specific hyperparameters are taken into consideration in this study, including the architecture of the variational function $q(\cdot)$ and its associated hyperparameters and the temperature of $q^{\sim}(\cdot)$. For ID-GFlowOut, there is an additional hyperparameter, which is the prior $p(z)$. The parameters are picked via grid search using the validation set. The temperature of $q^{\sim}(\cdot)$ is set as 2. In addition, with a 0.1 probability, the forward policy will choose a random mask set in each layer.

\subsubsection{Computational efficiency} On a single RTX8000 GPU, training models with GFlowOut takes around the same time as Contextual dropout and Concrete Dropout, and around twice the time (ResNet 7 hrs and MCAN Transformer 16 hrs) as a model with the same architecture and random dropout. The three learned dropout methods have similar efficiency during inference.

\subsection{Experimental details}

\subsubsection{Sampling dropout masks}
In the forward pass during inference, 20 samples are used for each data point. In ResNet experiments, dropout masks are generated for each ResNet block. In the transformer VQA experiment, in each layer, dropout is applied to both the self-attention and the feed-forward layer.

\subsubsection{Robustness to distribution shift}
The performance of each method was obtained with 9 repeats of different random seeds for training. Early stop using validation set was used to prevent overfitting. VQA Transformer experiments are designed according to \citet{yu2019deep}. 

\subsubsection{OOD detection}
For each data point, we take 20 forward passes and calculate the Uncertainty for prediction on each example using the Dempster-Shafer metric \cite{sensoy2018evidential} and  algorithm from \citet{jain2021deup}. The uncertainty score is used for classification of in-distribution vs. out-of-distribution data points assuming the later should have higher uncertainty.

\subsubsection{Adaptation after training on noisy data}
When training the model with noisy CIFAR-10/100 data, randomly picked 30\% data points are assigned a random label in the whole training set. The model obtained is then fine-tuned using a small number of clean data points all with correct labels. We conducted experiments with 1000,2000,4000 and 8000 data points used for fine-tuning.

\subsubsection{Real-world clinical data}
The ICU dataset is a real-world dataset, containing information about the deaths or survival of 126489 patients, across 58 different hospitals, given a set of administrated drugs. The goal of this experiment is to evaluate how well our approach generalizes, in real-world settings. To imitate this, we built two sets:
\begin{itemize}
    \item a training set that contains data points about patients from all hospitals, except the hospital with the highest number of patients (hospital ID 167). This results in a dataset with 120945 entries, which is equally partitioned (70:30 ratio) into the real training and validation sets.
    \item a test set that contains information about 5544 patients. As each hospital follows a specific distribution, the test set was designed to measure the OOD efficiency of GFlowOut, on the widest possible set of patients, which is a real-world scenario. 
\end{itemize}
We used a 3-layer MLP with multiple Dropout options as presented in Table \ref{tab:clinical}. For the evaluation, we perform 20 forward passes and take the mean of the prediction.

\subsection{Analysing dropout masks} 
\begin{figure*}[]
    \centering

    \includegraphics[width=0.4\linewidth]{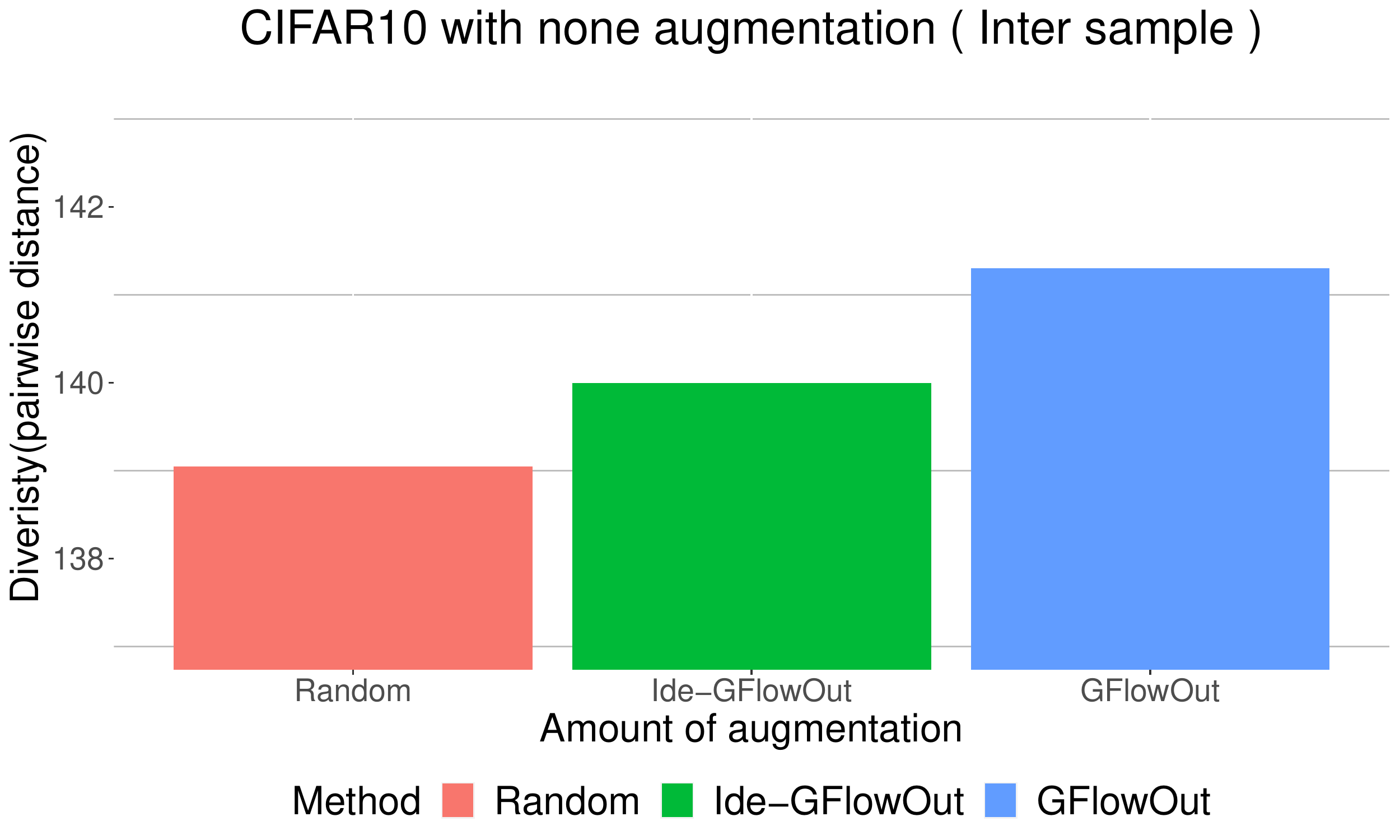}
   \includegraphics[width=0.4\linewidth]{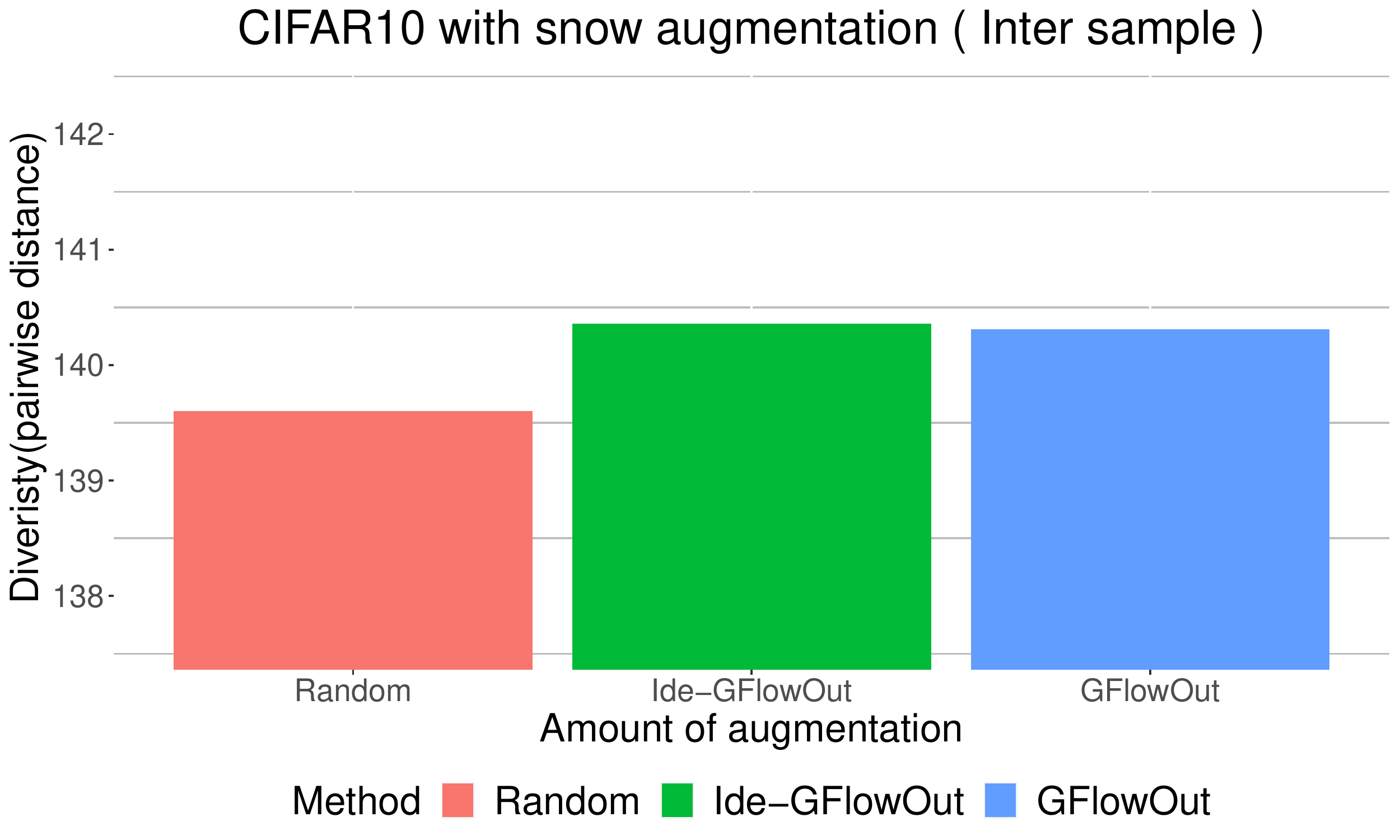}

    \includegraphics[width=0.4\linewidth]{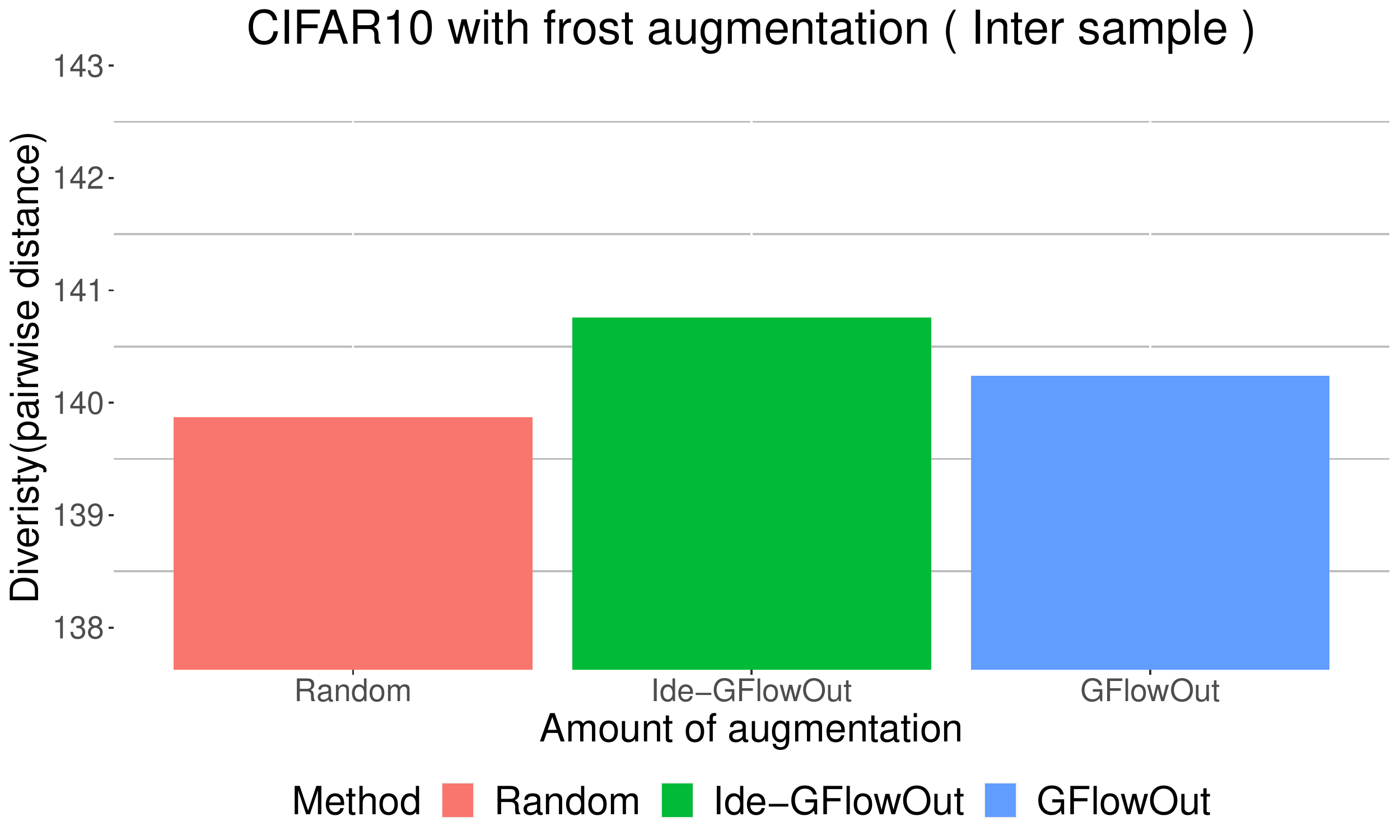}
    \includegraphics[width=0.4\linewidth]{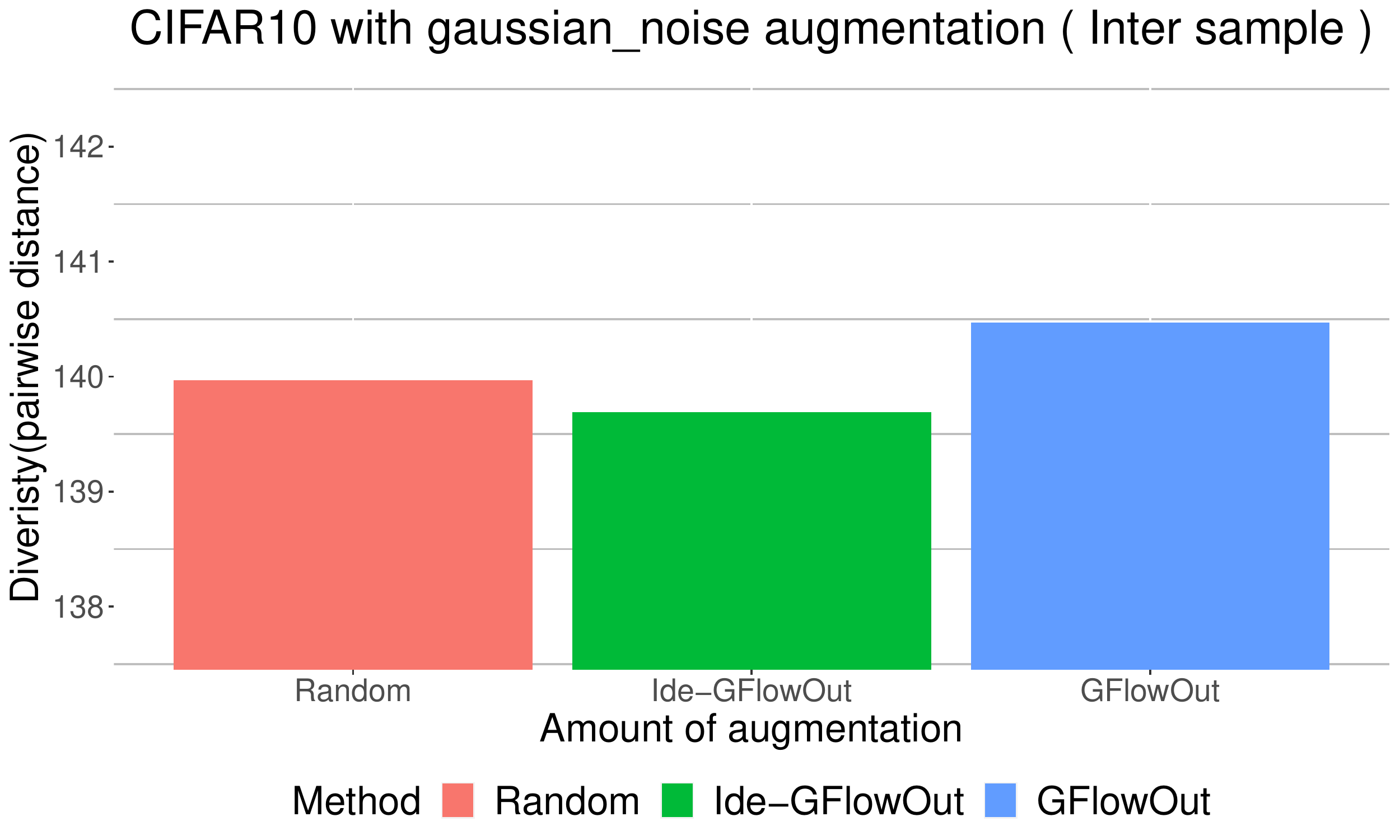}

    \caption{Diversity of binary dropout masks among different data points measured by Manhattan distance}
    \label{fig:Diveristy}
\end{figure*}

Here, we analyze the behavior and dynamics of the binary masks generated by GFlowOut for data points corresponding to different labels and different augmentations. First, we want to verify that GFlowOut generates masks with probability proportional to the reward $R$ as defined in equation (7). Our analysis shows a statistically significant correlation between the probabilities of a set of masks being generated by GFlowNet and the corresponding rewards, with correlation $\geq$ 0.4 and p values $\leq$ 0.05. Next, we want to explore whether GFlowOut generates diverse dropout masks. We take the mean dropout masks generated during inference for each data point and calculate Manhattan distances among different samples in the data set. The results are shown in Figure \ref{fig:Diveristy}.

\begin{figure*}[]
    \centering

    \includegraphics[width=0.45\linewidth]{figures/CIFAR-10_all.pdf}
    \includegraphics[width=0.45\linewidth]{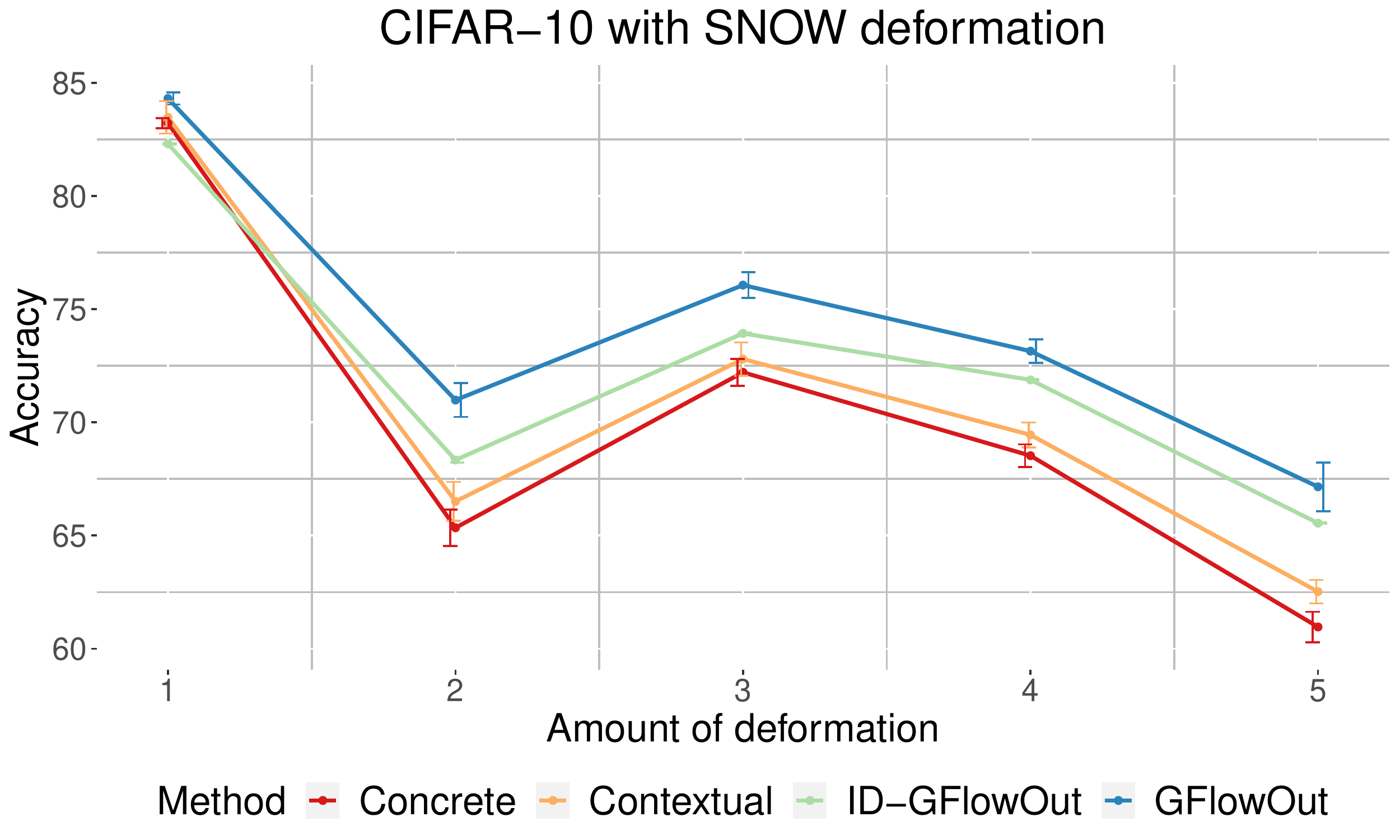}
    \includegraphics[width=0.45\linewidth]{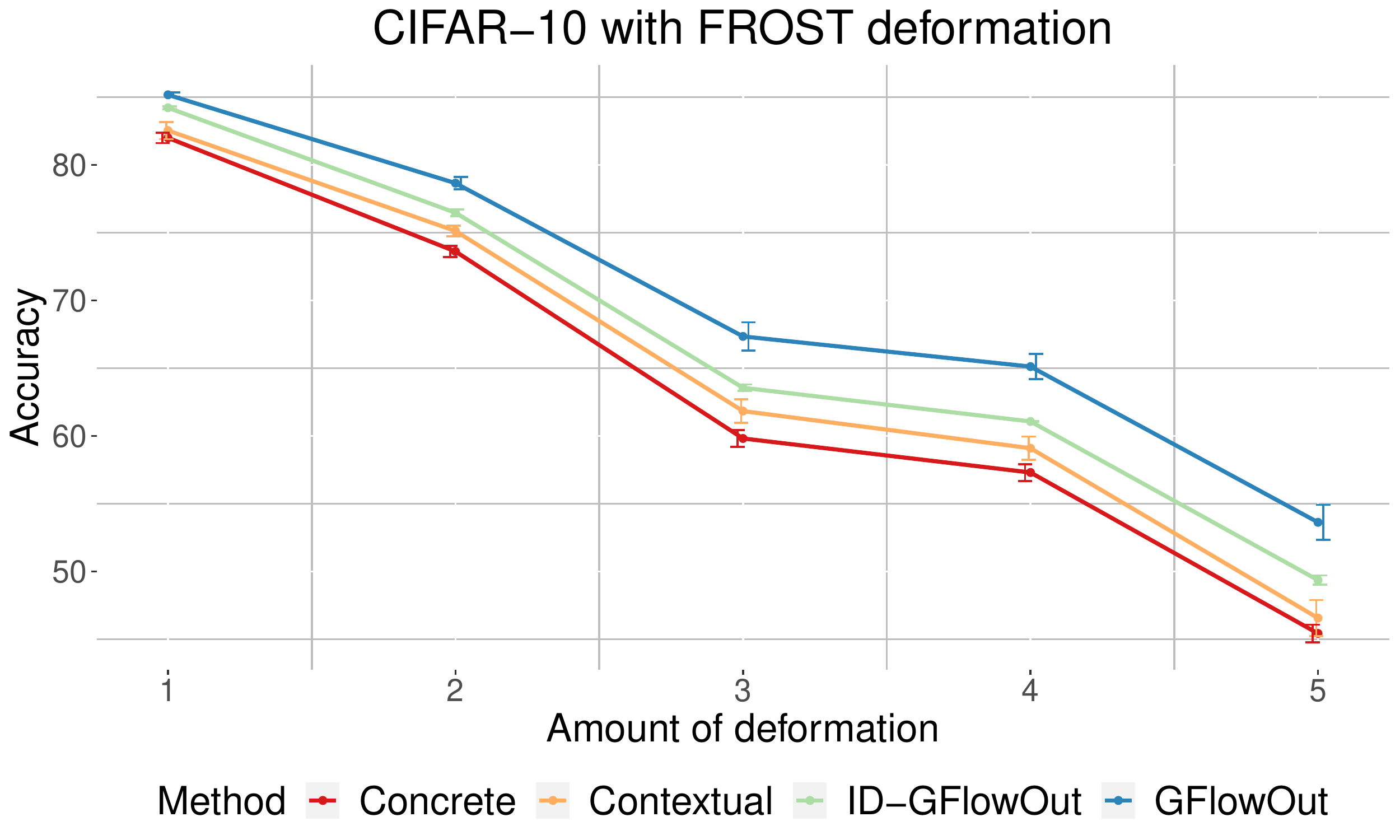} 
    \includegraphics[width=0.45\linewidth]{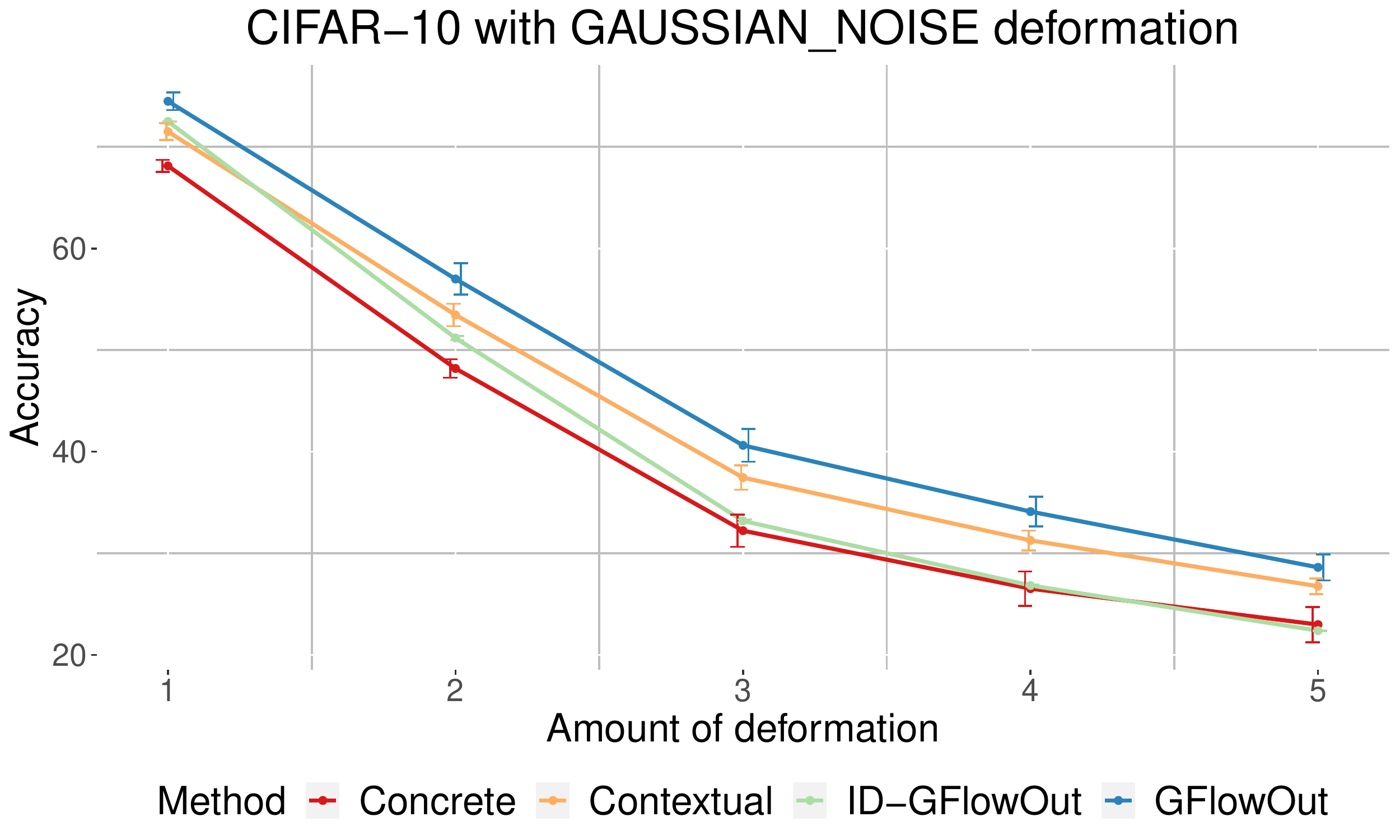}

    \caption{Robustness of ResNet-18 models trained  with different dropout methods to different amounts of SNOW deformation, FROST deformation or Gaussian noise  in  CIFAR-10 during test time}
    \label{augmentation_CIFAR-10}
 
\end{figure*}

\begin{figure*}[]
    \centering

    \includegraphics[width=0.45\linewidth]{figures/CIFAR-100_all.pdf}
    \includegraphics[width=0.45\linewidth]{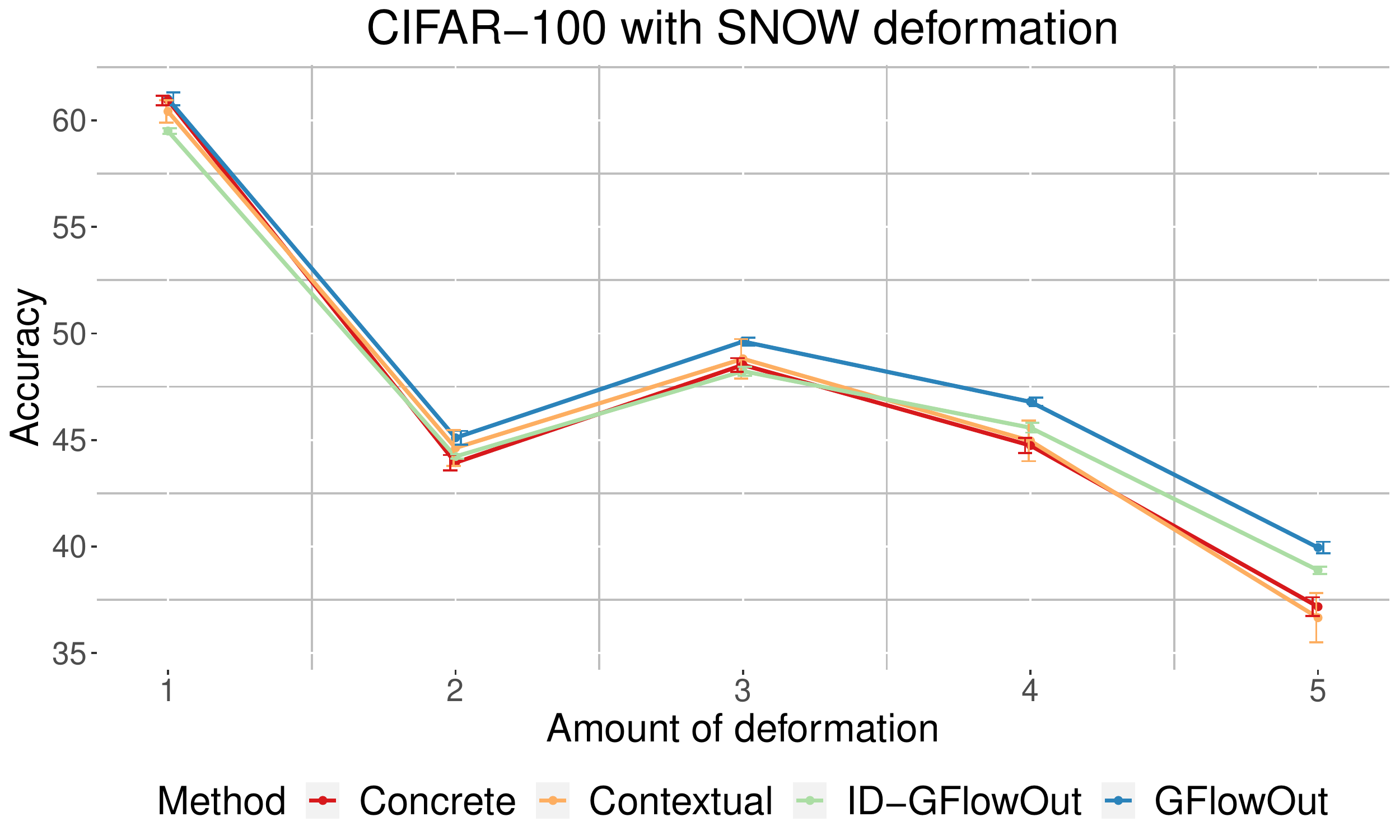}
    \includegraphics[width=0.45\linewidth]{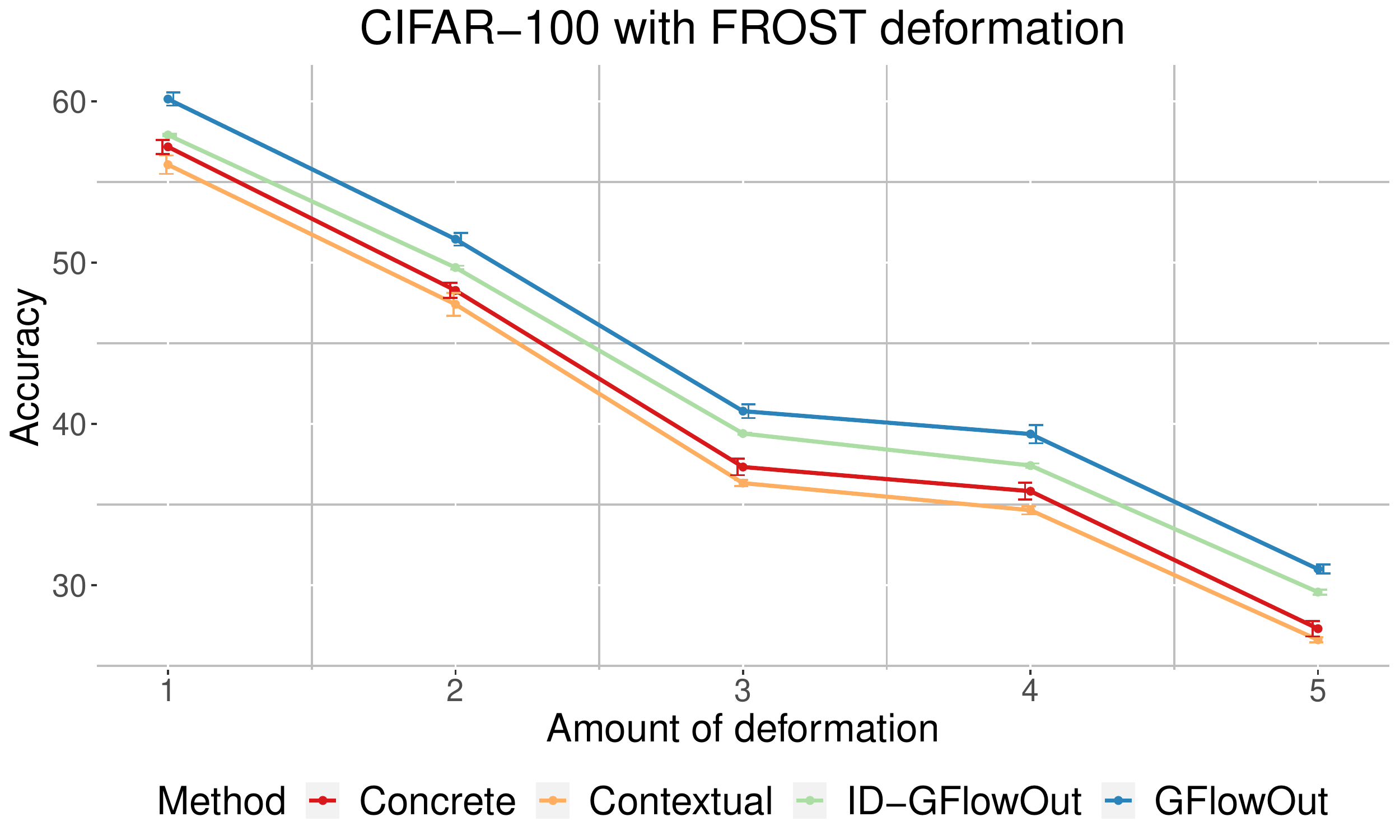} 
    \includegraphics[width=0.45\linewidth]{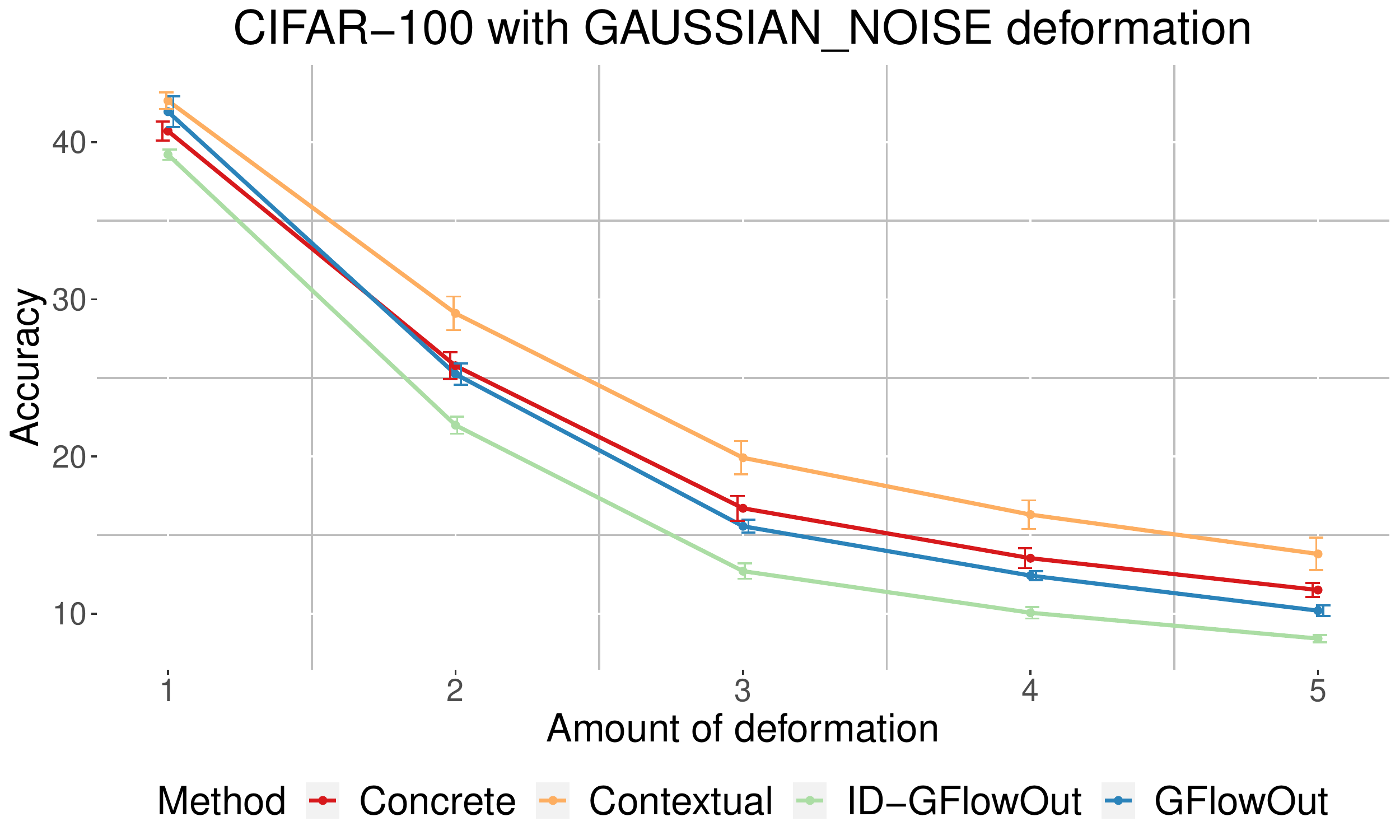}

    \caption{Robustness of ResNet-18 models trained  with different dropout methods to different amounts of SNOW deformation, FROST deformation or Gaussian noise  in CIFAR-100 during test time}
    \label{augmentation_CIFAR-100}
\end{figure*}

\begin{figure*}[]
    \centering

    \includegraphics[width=0.55\linewidth]{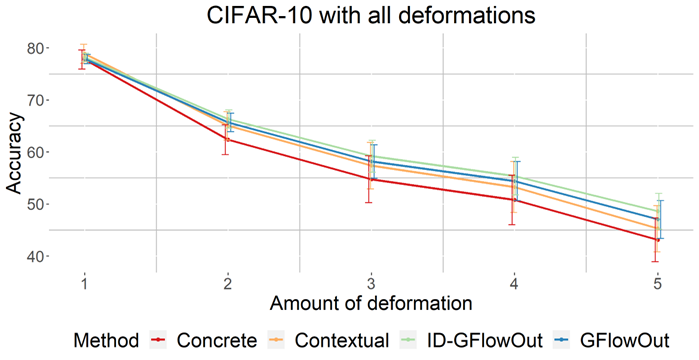}
    \includegraphics[width=0.55\linewidth]{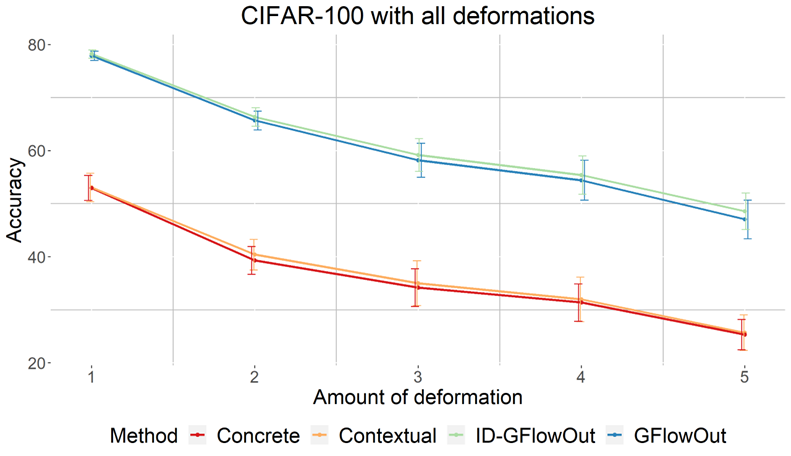}

    \caption{Robustness of ResNet-18 models trained with different dropout methods to different amounts of deformation using Resnet blocks with components in a slightly different order: BatchNorm -Conv-BatchNorm-Conv }
    \label{reorder_CIFAR}
\end{figure*}

\end{document}